\documentclass{article}
\usepackage[utf8]{inputenc}
\usepackage{graphicx}

\title{Machine Learning Methods for Management UAV Flocks - a Survey}
\author{Rina Azoulay, Yoram Haddad, Shulamit Reches\\
Jerusalem College of Technology\\
Jerusalem, Israel\\
email: \{azrina,haddad,reches\}@g.jct.ac.il}

\begin{document}
\maketitle

\begin{abstract}
The development of unmanned aerial vehicles  (UAVs) has been gaining momentum in recent years owing to technological advances and a significant reduction in their cost. UAV technology can be used in a wide range of domains, including communication, agriculture, security, and transportation.
It may be useful to group the UAVs into clusters/flocks in certain domains, and various challenges associated with UAV usage can be alleviated by clustering.
Several computational challenges arise in UAV flock management, which can be solved by using machine learning (ML)  methods. In this survey, we describe the basic terms relating to UAVS and modern ML methods, and we provide an overview of related tutorials and surveys. We subsequently consider the different challenges that appear in UAV flocks. For each issue, we survey several machine learning-based methods that have been suggested in the literature to handle the associated challenges. Thereafter, we describe various open issues in which ML can be applied to solve the different challenges of flocks, and we suggest means of using ML methods for this purpose.
This comprehensive review may be useful for both researchers and developers in providing a wide view of various aspects of state-of-the-art ML technologies that are applicable to flock management.

\end{abstract}

\section{Introduction}
\label{section:intro}

The technological advances in the abilities of unmanned aerial vehicles (UAVs) and the reduction in their costs have made UAVs more common than ever, with a significant potential for future use. During the Covid-19 period, UAVs could theoretically be used to assist in supply and transportation goals. Moreover, it is expected that UAVs will be integrated into the 5G and future communication networks. UAVs offer numerous applications in agriculture, maintenance, intelligence, and diverse security needs. However, several challenges should be overcome to leverage the potential of UAVs for practical goals successfully. For example, there is a need for a smart, efficient power control mechanism owing to the energy consumption limitation of UAVs, which are constrained in their weight-carrying ability. Moreover, aerial path planning should be performed carefully, while maintaining the security and safety of UAVs.

\begin{figure}
\includegraphics[width = 12cm]{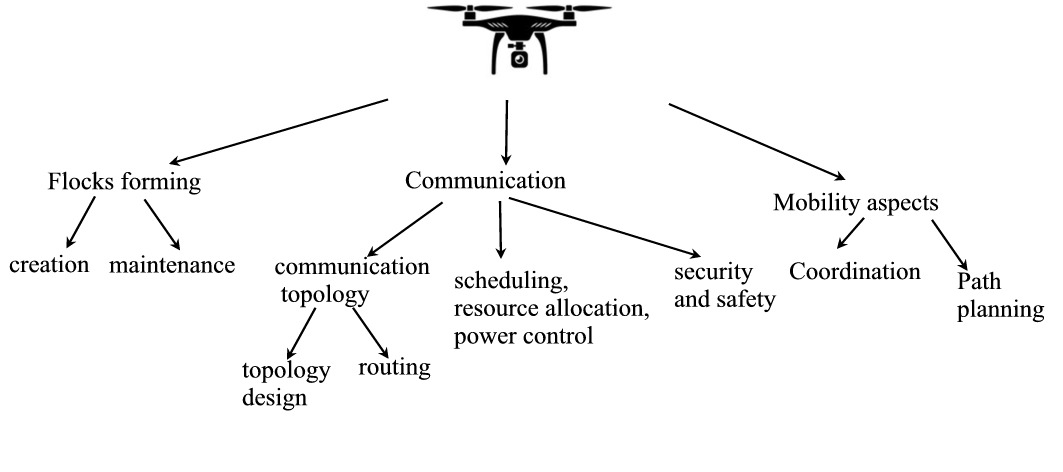}
   
\caption{Challenges in UAV flocking}
\label{figure:uav}
\end{figure}

In addition to the aforementioned challenges that occur even with a single UAV, certain tasks and issues arise when considering a multi-UAV environment, in which a small or a large group should act together or should act in the same aerial environment. For example, the formation of UAV networks, UAV clustering and coordination, and resource allocation in the UAV network should be considered and efficiently solved.
A hierarchical description of the various topics relating to UAVs is provided in Figure \ref{figure:uav}.

As part of this survey, we also provide several insights into general studies that are not directly focused on UAV environments, but that may be useful when applied to UAV environments. We believe that a complete view of the existing works, both those that have been specifically developed for UAVs and those that have not, can help UAV designers to address the challenges that arise in managing and maintaining UAV flocks more efficiently. 

As demonstrated in Section~\ref{section:surveys}, numerous existing surveys have dealt with the challenges and opportunities of UAV technologies. However, our goal is to focus on the increasingly important topic of applying machine learning (ML) methods to UAV flock domains. 

The remainder of this paper is organized as follows: Section~\ref{section:ML} provides basic terminology for modern ML methods that may be applicable to the improved handling of UAV flocks. Section~\ref{section:surveys} presents several relevant surveys that have considered UAV technologies and ML methods used for UAVs, as well as flying ad-hoc network (FANET) environments.  Section~\ref{section:formation} considers relevant studies that have dealt with flock formation. In Section~\ref{section:resource}, the challenges of resource allocation and power control are discussed, whereas Section~\ref{section:allocation} focuses on studies that have addressed the challenging problem of task allocation in UAV flocks.
Finally, Section~\ref{section:conclusion} presents several conclusions and insights into future works.


\section{Some Basic Terms and Background}
\label{section:ML}

\begin{figure}
\includegraphics[width = 12cm]{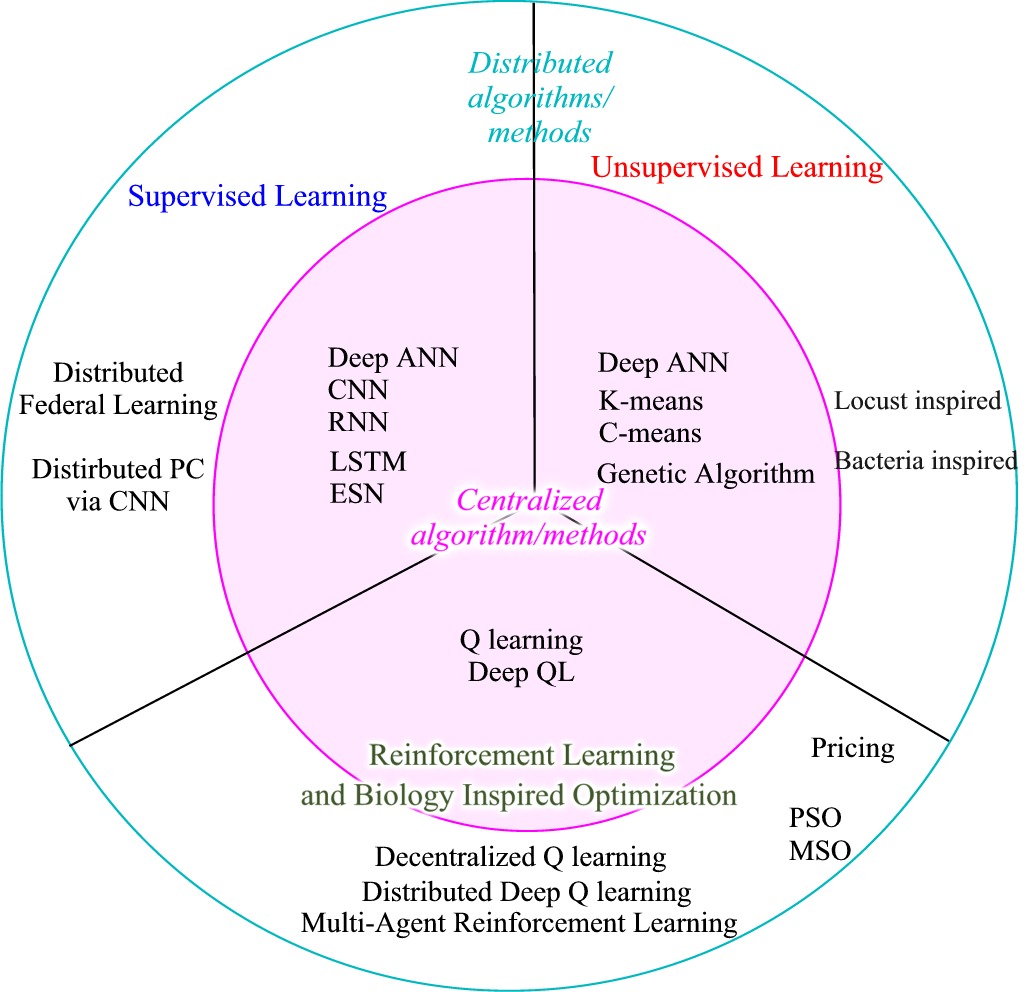}
   \caption{Main ML methods}
\label{figure:ML2}
\end{figure}

As the main goal of our survey is to review the application of ML methods to advance artificial aerial swarm scenarios, we first briefly present the main ML methods considered in the literature that are reviewed within the framework of this survey.
Figure~\ref{figure:ML2} presents a scheme  with the details of several ML methods reviewed in this survey, and In Table~\ref{table:ML:Methods}, we list the main ML methods described in this survey, their main features, and their applicability to various UAV flock challenges studied in this survey. 

\begin{table}
\caption{Main ML methods (C = centralized, D = distributed)}
\label{table:ML:Methods}
\begin{center}
  \begin{tabular}{ | l | l | l | l | l |}\hline
ML & Learning type & Approach  &  Main UAV     \\ 
 Method  &   &  &      Applications     \\ \hline 
 Deep learning  & Supervised/unsupervised  & C/D &           Resource allocation \\ \hline 
 CNN  & Supervised  & C     &  Clustering, resource allocation     \\ \hline 
 RNN  & Unsupervised  & C     & Task allocation       \\ \hline 
 LSTM  & Supervised  & C     &  Deployment     \\  \hline
 Federal learning  & Supervised &  C/D    & Resource allocation       \\ \hline 
 Q-learning  &  Reinforcement learning  & C/D &  Clustering, flocking,         \\ 
 Reinforcement learning  &     &   &    Deployment, positioning,         \\ 
      &     &   &    Resource allocation         \\ 
   
   \hline 
Deep Q-learning,   &   Reinforcement learning  & C/D & Flocking, deployment, \\
Deep Reinforcement learning   &     &   &    positioning, resource        \\    &     &   &    allocation, task allocation      \\ \hline 
Particle swarm optimization  &   Bio-inspired & D & Clustering, routing  \\  
 &    &  &  positioning, task allocation        \\ \hline 
Genetic algorithm    & Bio-inspired & C & Clustering, deployment,      \\ 
     &   &   &   task allocation      \\ \hline 
K-means & Unsupervised  &  C    & Clustering, deployment,        \\ 
  &    &       &  resource allocation,        \\ 
  &    &       &  task allocation      \\ \hline 
Pricing & Unsupervised & D & Resource allocation  \\ \hline
   \end{tabular}
\end{center}
\end{table}

Various UAV types were used in the reviewed papers.
An overview of UAV types can be found in \cite{HentatiKrichen}. The authors provided a detailed description of the available simulation tools for UAV system performance, including the requirements,
goals, and strengths and weaknesses of each tool. They divided the UAVs into three categories: micro UAVs (the weight of which can reach 2 kg), mini UAVs (2 to 20 kg), and small UAVs (20 to 150 kg).
Micro UAVs have an endurance of several hours, and can be used for reconnaissance and inspection, whereas mini and small UAVs can last for up to two days and can be used for data gathering. All UAV types can be used for various surveillance goals.

Regarding UAV flocks, Coppola et al. \cite{Coppola} distinguished among different flock types as follows:
\textbf{Teams} are typically small groups, where each agent optimizes individual objectives in a cooperative or competitive manner.
\textbf{Formation} refers to a situation where each agent is typically assigned a specific sub-task, role or placement
in a cooperative manner, whereas \textbf{swarms} typically refer to large groups of dispensable agents. Nevertheless, in this survey, we consider all types of UAV 
formations as \textbf{flocks}. 

In Table~\ref{acronyms}, we provide a list of acronyms used throughout this paper.
\begin{table}
\caption{List of acronyms}
\label{acronyms}
\begin{center}
\begin{small}
   \begin{tabular}{ | l | l |}\hline
Acronym & Text      \\  \hline
ABS & Aerial base station \\
ACO & Ant colony optimization \\
AI & Artificial intelligence \\
ANN & Artificial neural network \\
APC & Affinity propagation clustering \\
AQL & Adaptive Q-learning \\
BS & Base station \\
CH & Cluster head \\
CMTAP & Cooperative multiple task assignment problem\\
CNN & Convolutional neural network \\
CPS & Cyber-physical system \\
DNN & Deep neural network \\
DRL & Deep RL \\
ESN & Echo state network  \\
FANET & Flying ad-hoc Network \\
FL & Federal learning \\
FTA & Fast task allocation \\
GA &  Genetic algorithm \\
GPI & Geographical position information \\
HAPS & High-altitude platform station \\ 
HTER & Half total error rate \\
IoD & Internet of drones \\
LEACH & Low-energy adaptive clustering hierarchy \\
LEACH-C & LEACH-centralized \\
LP & Linear programming \\
LSTM & Long short-term memory \\
MA & Multi-agent \\
MADDPG & Multi-agent deep deterministic policy gradient \\
MARL & Multi-agent reinforcement learning \\
MAV & Micro air vehicle   \\
MDP & Markov decision process  \\ 
MEC & Mobile edge computing \\
MFG & Mean-field game \\
ML & Machine learning  \\
NDP & Neuro-dynamic programming \\
NFV & Network function virtualization \\
NLP & Nonlinear programming \\
PSO & Particle swarm optimization \\
QoE & Quality of experience \\
RAN & Radio access network \\
RL & Reinforcement learning \\
RPW & Red palm weevil \\
SDN & Software-defined network \\
SI & Swarm intelligence \\
SPNE & Subgame perfect Nash equilibrium \\
STAPP & Simultaneous target assignment and path planning \\
UAV & Unmanned aerial vehicle  \\
WSN & Wireless sensor network \\
\hline

\end{tabular}
\end{small}
\end{center}
\end{table}

\section{Recent Tutorials and Surveys}
\label{section:surveys}  


The aim of this study is to review recent works that have considered the concept of applying state-of-the-art ML methods to handle the challenges of UAV flocks. In recent years, several tutorials and surveys have been presented on related topics, such as the application of ML methods to solve challenges in UAV design and management, or surveys dealing with flocking challenges. However, no previous surveys have focused on the use of ML for solving UAV flocking challenges. \\
Therefore, in this section, we first review existing surveys and tutorials that are closely related to our goal. We categorize these various studies into two main categories. Section~\ref{section:flocks} focuses on the challenges in flock and swarm creation and management, including several flocking issues such as efficient deployment and security enhancement, whereas Section~\ref{section:fanet} focuses on surveys that have considered using ML techniques for enhancing UAV technologies and the communication between UAVs. Furthermore, Section~\ref{survey:ML} discusses studies that have focused on using ML methods in this area, and finally,
several comparisons and a generalization of the above.

\subsection{UAV Flocks Creation, Management and Coordination}
\label{section:flocks}

The creation of groups of UAVs is important in several applications, such as wireless communication applications, agriculture applications, and security purposes. In the following section, we review several surveys that have dealt with challenges considering flock creation and management, where the flocks can be used for several application types.
The significant advancements in UAV technology have resulted in applications whereby groups of UAVS (which are known as flocks or swarms) can cooperate to perform a global task. This may be the case when considering UAVs that act as cellular base stations (BSs),  which should decide among themselves how to offer deployment to provide good coverage of an area for ground users. Other scenarios encompass different domains where UAVs can be used, such as photography tasks, control tasks, transportation applications, security purposes, and agriculture domains. The common factor of all these applications is that they gave rise to several challenges whereby decisions should be made, such as UAV deployment, task allocation, and maintenance issues. In the following, we describe several surveys relating to the aforementioned issues, some of which also discuss relevant ML solutions to the problems reviewed.\\
\\

The significant potential of the general use of UAVs, and more specifically, UAV flocks, was mentioned by Tahir et al. \cite{tahir}. The authors analyzed the core characteristics of swarming drones, including the UAV mechanics, functionality, organization, modeling, and applications, as well as the autonomous aspects of such drones and drone swarms. Moreover, they conducted an online survey to determine the public awareness regarding the potential of using drone technology. The survey results demonstrated that, although most people were concerned about UAV usage, the majority did not know about the concept of a swarm of drones.

Although there is significant potential, several challenges also need to be overcome. 
Chung et al. \cite{ChungSurvey} surveyed 
algorithms that allow the individual members of the swarm to communicate and to allocate tasks among themselves, plan their trajectories, and coordinate their flight so that the overall objectives of the swarm can be achieved efficiently.

The authors reviewed several theoretical tools, with a specific focus on how they have been developed for and applied to aerial swarms, including theoretical algorithms, ML methods, and other advanced planning and decision-making strategies. They emphasized that an important area of further study is the development of learning and decision-making architectures that will endow swarms of aerial robots with high levels of autonomy and flexibility.

%
Coppola et al. \cite{Coppola} focused on micro air vehicles (MAVs), which are miniature UAVs with a size as small as 5 cm, and reviewed the challenges that must be addressed to develop MAVs for real-world operations successfully. The challenges begin at the lowest level, in terms of how the MAV design will impact the swarm behavior, and end at the highest level, where collective behaviors must be designed that can best exploit the lower-level designs, controllers, and sensors.
Although the work of Coppola et al. did not focus on ML, they discussed several ML methods for behavior design and optimization. In particular, they investigated evolutionary robotics and reinforcement learning (RL).

Returning to the general problem of UAV flock formation, Beaver et al. \cite{beaver2020overview} presented an overview of optimization approaches for flocking behavior that provide strong safety guarantees. They summarized 
the results of existing optimal flocking works across engineering disciplines and presented the frontier of flocking and optimization research. They also aimed to provide a new paradigm for understanding flocking as an emergent phenomenon to be controlled, rather than desirable group behavior for agents to mimic.

cite{Sargolzaei} explored several applications of cooperative
UAV missions in various fields, and
reviewed emerging topics in the field of cooperative UAV control
and their associated practical approaches. They surveyed the applications, algorithms, and challenges. They categorized the applications
into surveillance, search and rescue, mapping, and military applications. Similarly, they categorized the algorithms into three main classes: consensus control, flocking control, and guidance-based cooperative control. Furthermore, they investigated the challenges relating to the cooperative control and applications of cooperative algorithms and provided the related mathematics of the cooperative control algorithms.

Several studies have drawn inspiration from the behavior of flocks in nature and used bio-inspired algorithms for the formation and management of UAV flocks/swarms.
Long et al. \cite{Long}

surveyed the literature on swarm shepherding, which is a method for controlling multiple coordinated robotic agents. 
The shepherding problem can be
defined as the guidance of a swarm of agents from an initial
location to a target location. The shepherd is
responsible for the high-level path planning of the swarm and
task allocation, whereas the swarm members themselves function
solely on single-agent dynamics.
Long et al. mentioned the use of ML and computational intelligence techniques for swarm control, and for shepherding in particular.

All the above surveys focused on defining general flock types and their management. Several also mentioned ML methods, but none concentrated on surveying ML methods for the creation and management of UAV flocks.

subsection{Surveys on UAV-Based Wireless Communication Networks}
\label{section:fanet}

The incredible technological advancements in the capabilities for organizing groups of drones have considerably expanded their use as part of the 5G and beyond systems.
A UAV-based wireless network is often referred to as a FANET. This term was formally defined by Bekmezci et al. \cite{Bekmezci}, who suggested several applications for FANET and discussed various design challenges. FANET design considerations were investigated, including the adaptability, scalability, latency, UAV platform constraints, and bandwidth.

Hayat et al. \cite{Hayat16} discussed the characteristics and requirements of UAV networks for envisioned civil applications over the period of 2000 to 2015 from a communications and networking perspective. They surveyed and quantified the quality-of-service requirements, network-relevant mission parameters, data requirements, and minimum data to be transmitted over the network. Furthermore, they elaborated on the general networking-related requirements, such as connectivity, adaptability, safety, privacy, security, and scalability. They reported experimental results from many projects and investigated the suitability of existing communication technologies for supporting reliable aerial networking.

Recent implementations and projects using multi-UAV systems were surveyed by Hentati et al. \cite{Idriss}.
They presented a review of the recent UAV communication protocols and mechanisms and summarized current research in the area of monitoring via UAVs.
They discussed well-known design issues relating to UAV communication protocol, including mobility, adaptability, reliability, scalability, latency, UAV platform constraints, bandwidth allocation, power consumption, computational power, localization, security, and privacy. Moreover, they surveyed existing multi-UAV projects, testbeds, and simulation environments.
Another recent study by 

Boccadoro et al. \cite{Boccadoro}
 categorized the multifaceted aspects of the Internet of drones (IoD),
 including several fields in which drones and swarms of drones may be useful. Their goal was to provide an overview of the current state of research activities regarding the IoD. In particular, they proposed a classification scheme that follows the Internet protocol stack, starting from the physical layer and moving up towards the application layer, without
neglecting cross-layer approaches.

The survey of 
Fotouhi et al. \cite{Fotouhi} focused on regulation issues that arise when integrating UAVs into cellular systems. They reviewed the available UAV types and potential solutions to interference issues, as well as challenges and opportunities for assisting cellular communications with UAV-based flying relays and BSs. They also considered testing, regulation, and security challenges.

Oubbati et al. \cite{Oubbati2} summarized
the main characteristics of software-defined network (SDN)
and network function virtualization (NFV) technologies that are used for the efficient management of UAV-assisted mobile networks.
They presented an
in-depth discussion relating to both the design challenges of
UAV networks and their principal use cases. Moreover, they provided an overview of the SDN and NFV concepts and how they can be seamlessly integrated into UAV networks, and reviewed the majority of recent research activities on SDN/NFV-enabled UAV networks based on different use cases, such as UAV-assisted cellular communications, routing, monitoring, security, and several
other applications.

The authors emphasized that AI techniques and ML methods are expected to play a crucial role in
optimizing UAV-assisted networks in various aspects, such
as optimizing the resource allocation and scheduling,
enhancing the network prediction and boosting the network performance. However, multiple challenges require further investigation, such as the high computational processing, high energy
consumption, and high latency.


A special focus on the algorithmic challenges involved in UAV-based networks was provided in the study of
Ullah et al. \cite{Ullaha}. They investigated the joint optimization problem to enhance the system efficiency of UAV-assisted next-generation communication systems.
They focused on the following challenges: flight trajectory, time scheduling, altitude optimization, aerial and relay BSs, energy harvesting, power transfer, optimal power consumption, and
resource allocation. They presented an in-depth study regarding various successive optimization algorithms and methods, such as mobile edge computing (MEC) techniques and SDNs, as well as ML and
deep learning applications that play a vital role in 5G and B5G
communication.

The potential of using ML methods for UAV-based communication networks has been the highlight of several recent surveys.
Mozaffari et al. \cite{Mozaffari} surveyed the applications and challenges of UAV usage in communication networks, as aerial BSs (ABSs), or as flying mobile terminals. Specifically, they also noted issues to which ML can be applied, namely channel modeling and trajectory optimization.

Saad et al. \cite{saad} surveyed 
the countless applications of UAVs and UAV swarms, and the challenges that
should be addressed when considering these technologies. They reviewed research problems pertaining to UAV network performance analysis and optimization, including the physical layer design, trajectory path planning, resource management, multiple access, cooperative communications, standardization, control, deployment strategies, and security issues.
Sharma et al. \cite{SharmaSurvey} provided insights into the latest UAV communication technologies through the investigation of suitable task modules, antennas, resource handling platforms, and network architectures. They considered the applicability of ML, encryption, and optimization techniques to ensure long-lasting and secure UAV communications.

Moreover, several surveys have mainly focused on describing the major ML methods and their application to UAV wireless communication networks, which we review in Section~\ref{survey:ML}.

Owing to the common use of skimmers in security areas, one of the most challenging issues in the implementation of a wireless UAV-based network is the need for protection and security.

Shakeri et al. \cite{Shakeri30}
surveyed several design challenges
of multi-UAV systems for cyber-physical system (CPS) applications. They considered the modern generation of systems with synergic cooperation between computational and physical
potentials that can interact with humans through several new
mechanisms. UAVs can act as part of such systems, and the potential of ML to improve the efficiency of CPSs was discussed.

  Wang et al. \cite{WangZhao} presented a comprehensive survey on UAV networks from a CPS perspective. They attempted to provide novel insight into the state of the art in UAV networks. The basics of the three cyber components, namely communication, computation, and control, were discussed, including the requirements, challenges, solutions, and advances. The three UAV network hierarchies, namely the cell level, system level, and system of the system level, were investigated according to the CPS scale, with the architecture, key techniques, and applications of the UAV network explicitly demonstrated under each hierarchy. Finally, the coupling effects and mutual inspirations among the UAV network components were explored, which will likely lead to solutions that address the challenges from a cross-disciplinary perspective.
Blockchain applications in UAV networks have even been considered in \cite{AlladiBlockchain}. Such applications include network security, decentralized storage, inventory management, and surveillance. Broader perspectives were also discussed. They presented various challenges to be addressed in the integration of blockchain and UAVs and suggested several future research directions. It worth noting that they also considered ML as one of the blockchain applications in other emerging areas.

%
As the focus of our survey is on the use of UAV flocks/swarms, we present the following surveys relating to the use of UAV swarms in wireless communication networks.
Li et al. 
 \cite{LiFeiZhang} presented space–air–ground integrated networks and discussed the research challenges faced by the emerging integrated network architecture. They also provided a review of various 5G techniques based on UAV platforms that could be categorized into different domains, such as the physical layer and network layer, as well as joint communication, computing, and caching. They defined a UAV-based swarm network as a network based on a swarm of UAVs that offers ubiquitous connectivity to ground users. The benefits of such a network are that it is highly flexible with rapid provision features, and it is a feasible solution to enable fast, effective recovery and communication expending.

Another survey relating to UAVs for wireless networks was conducted by Zhang et al. \cite{ZhangLi}. They
discussed several aspects of using air–ground integrated mobile edge networks, in which UAVs are employed to assist the MEC network.
Among the advantages of this approach, they provided the following examples: the easy deployment of such networks, the high mobility of UAVs, and the ability to provide
MEC services ubiquitously and reliably.
They also considered the advantages as well as the challenges of using UAV swarms, which can complete tasks more efficiently and can improve the chance of successful missions. Furthermore, they can improve the scalability and reduce the probability of system failure. However, they noted that the means of hiring UAVs requires further discussion.

The above survey focused on the use of UAVs as part of communication networks. However, as mentioned previously, UAVs have several purposes, and somehow, different challenges arise when considering a communication network from when aiming to connect the UAVs among themselves.
XiJun et al. \cite{XiJun} focused on UAV swarm communication architectures and routing protocols.
They detailed four communication architectures for UAV swarms. They also discussed applicable scenarios and introduced a systematic overview of as well as feasibility research on routing protocols. Finally, they investigated the open research issues of UAV swarm communication architectures and routing protocols. However, they hardly addressed ML methods.

In the following section, we present surveys that have focused on ML methods used for solving different challenges arising in UAV technologies and UAV swarms.

\subsection{Surveys on Application of ML to UAV-Related Challenges}
\label{survey:ML}

In the following, we mainly focus on ML methods used for UAV control and UAV flock control.
Considering a group of UAVs in terms of a communication network offers several advantages as well as faces various challenges. In fact,
general-purpose UAVs can form a D2D wireless network among themselves in order to be coordinated. In other applications, the mesh network of the UAVs can be used as a cloud that provides a wireless network for ground users.

Bithas et al. \cite{Bithas} presented a survey on the use of ML techniques for UAV-based communication and the improvement of various design and functional aspects. Their survey focused on UAVs that are used for communication tasks, as part of 5G and beyond communication systems, in which the following aspects were detailed: channel modeling, interference management, parameter configuration, security and safety, resource management, position detection, and localization. However, our survey differs from the study of Bithas et al. in terms of the following aspects: (a) we concentrate on the challenges relating to the planning, organization, and maintenance of UAVs flocks, rather than dealing with the challenges of a single UAV; (b) we provide a brief description of classical and innovative ML methods and their use in the construction, management, and utilization of UAV flocks for different goals; and (c) we provide an overall survey of related studies that may be applicable for these challenges, even if they did not directly consider the UAV flocking goal.

Several studies have reviewed the capability of artificial neural networks (ANNs) and advanced deep learning methods to address some of the above challenges relating to UAV applications.
Carrio et al. \cite{Carrio} reviewed several applications of deep learning for UAVs, including the most relevant developments as well as their performances and limitations.

Mao et al. \cite{Mao}
performed a comprehensive survey of the applications
of deep learning algorithms for different network layers, including physical
layer modulation/coding, data link layer access control, resource
allocation,  and routing layer path searching and traffic balancing.
They also discussed the use of deep learning to enhance other network functions, such as network security and sensing data compression.

Chen et al. \cite{Chen} considered the potential of using ANNs for solving various wireless networking problems. They
presented a detailed overview of several recent ANN types, including recurrent, spiking, and deep neural networks,
which are pertinent to wireless networking applications. For each
ANN type, they presented the basic architecture as well as specific
examples that are particularly important for wireless network
design. Moreover, they provided
an in-depth overview of the variety of wireless communication
problems that can be addressed using ANNs. For each individual application, they presented the main motivation for using ANNs, along with the associated challenges
and use cases.

Another survey considering deep learning for wireless networking was presented by Zhang et al. \cite{Zhang2}.
They introduced and compared state-of-the-art deep learning models, and provided guidelines for model selection towards solving networking problems, whereby the various scenarios and applications in wireless networking were grouped, ranging from mobile traffic analytics to security and emerging
applications.

The recent study of \cite{Kouhdaragh} examined how ML can be used in the area of radio access networks (RANs) with UAVs. The paper investigated which types of ML methods are useful for designing UAV-based RANs (U-RANs), with a particular focus on supervised RL strategies.
They discussed the advantages and potential of using ML algorithms in U-RANs in numerous network design scenarios that could not be easily solved by conventional model-based methods. They investigated the different types of ML methods that are useful for designing U-RANs by focusing on supervised and RL strategies in particular.


Al-Turjman et al. \cite{AlTurjman}
presented an overview of the use of artificial intelligence (AI) in UAV communications. They reviewed the communication protocols and technologies that have been applied in UAV communications and discussed several AI-based classification and image-based techniques that are used for UAV communication.
Another recent study on UAV-based communication with ML method is the study of Wu et al. \cite{Wu5G}, which focused on the exploitation of advanced techniques, such as
intelligent reflecting surfaces, short packet transmission, energy
harvesting, joint communication and radar sensing, and edge
intelligence, to meet the diversified service requirements of future wireless systems.
They also reviewed ML methods specifically for UAV
trajectory and communication design. Interestingly, Wu et al. considered the three research dimensions that are the focus of our survey, namely UAV technology, swarm control, and ML utilization. However, Wu et al. focused on advanced technologies that are used for UAV-based communication networks, whereas our study focuses on ML methods that are used for multipurpose UAV flocks and swarms.

The majority of the above surveys focused on the utilization of ML methods for solving FANET challenges, but none of them considered the challenges that are specific to UAV clustering, which is an important issue in several practical areas. Table~\ref{table:survey} summarizes the afore-mentioned reviews, with a special mention of the main focus of each one. In particular, for each survey, we note whether it focused on UAV/aerial vehicles, implicitly considered flocks and flock formation, and whether ML methods and applications were also surveyed. In the table, the notation "V" indicates a main concern, whereas "+" denotes that several relevant ideas were mentioned, but with no main focus on this issue, and finally, $\times$ means that the study (almost) did not consider the mentioned issue.

To summarize, Table~\ref{table:survey} presents a list of surveys related to the topic of UAVs flocks, and the challenges relating to their design and management.
\begin{table}
\caption{Summary of scope of surveys}
\label{table:survey}
\begin{center}
\begin{tabular}{ | l | l | l | l | l | }\hline
{\bf Reference} &   {\bf Scope} & {\bf UAVs} & {\bf Flocks } & {\bf ML }  \\ \hline
 Tahir et al.\cite{tahir} & Future usage of UAVs and flocks & $V$ & $V$ & $\times$ \\ \hline
Chung et al. \cite{ChungSurvey}& Aerial swarm robotics & $V$ & $V$ & $+$\\ \hline
Coppola et al. \cite{Coppola} & Challenges of Micro Air Vehicles & $V$ & $\times$ & $+$ \\ \hline
 Beaver et al. \cite{beaver2020overview} &  Optimal robotic flocking & $+$ & $V$ & $+$\\ \hline
 Sargolzaei et al. \cite{Sargolzaei} & Flock cooperation control& $V$ & $V$ & $+$\\ \hline
 Long et al. \cite{Long} & Swarm control & $\times$ & $V$& $V$ \\ \hline \hline
 Bekmezci et al. \cite{Bekmezci} & Flying ad-hoc networks & $V$ & $V$& $\times$ \\ \hline
 Hayat et al. \cite{Hayat16} & Civil applications & $V$ & $V$ & $\times$\\ \hline
 Hentati et al.\cite{Idriss}& Design issues & $V$ & $V$ & $+$ \\ \hline
 Boccadoro et al. \cite{Boccadoro} & Internet of Drones & $V$ & $+$ & $\times$\\ \hline
 Fotouhi et al. \cite{Fotouhi} & Standards of UAV based communications & $V$ & $\times$ & $+$ \\ \hline
 Ullah et al. \cite{Ullaha}& Optimizing  communication efficiency & $V$ & $+$ & $V$\\ \hline
 Mozaffari et al. \cite{Mozaffari}& UAV based communication & $V$ & $\times$& $+$\\ \hline
 Saad et al. \cite{saad}& UAV based networks & $V$ & $+$ & $V$ \\ \hline
  Sharma et al. \cite{SharmaSurvey}& Networking technologies & $V$ & $\times$ & $+$ \\ \hline
 Shakeri et al. \cite{Shakeri30}& Cyber-physical applications & $V$  & $V$ & $V$\\ \hline
 Wang et al. \cite{WangZhao}& Cyber UAV perspective & $V$ & $V$ & $V$ \\ \hline
Alladi et al. \cite{AlladiBlockchain}& Blockchain applications & $V$ & $\times$ & $+$\\ \hline
Li et al.  \cite{LiFeiZhang}& UAV communication & $V$ & $+$ & $V$ \\ \hline
Oubbati et al. \cite{Oubbati} & SDN and NFV for UAV assisted networks & $V$ & $V$ & $+$ \\ \hline
 Zhang et al. \cite{ZhangLi}& Air-Ground Integrated Networks & $V$ & $+$ & $+$\\ \hline
 XiJun et al. \cite{XiJun}& UAV swarm communication & $V$ & $V$ & $+$\\ \hline \hline
 Bithas et al. \cite{Bithas}& ML for UAV communication systems& $V$& $\times$ & $V$\\ \hline
 Carrio et al. \cite{Carrio}& Deep Learning for UAVs & $V$ & $\times$ & $V$ \\ \hline
 Mao et al. \cite{Mao}& Deep Learning for Wireless networks& $+$ & $+$ & $V$\\ \hline
 Chen et al. \cite{Chen}& ANNs for wireless networks & $V$ & $\times$ & $V$\\ \hline
 Zhang et al. \cite{Zhang2}& Deep Learning for wireless networks & $\times$ & $+$ & $V$ \\ \hline
 Kouhdaragh et al. \cite{Kouhdaragh}& ML for UAV-Based Networks Design & $V$ & $\times$ & $V$\\ \hline
 Alsami et al. \cite{Alsamhi32}& AI for robotic communication & $V$ & $V$ & $V$ \\ \hline
 Al-Turjman et al. \cite{AlTurjman}& AI-based UAV communication system & $V$ & $\times$ & $V$ \\ \hline 
 Wu et al. \cite{Wu5G}& Sensing and Intelligence & $V$ & $V$ & $V$  \\  \hline
  \end{tabular}
\end{center}
\end{table}

\begin{figure}
\begin{center}
\includegraphics[width = 10cm]{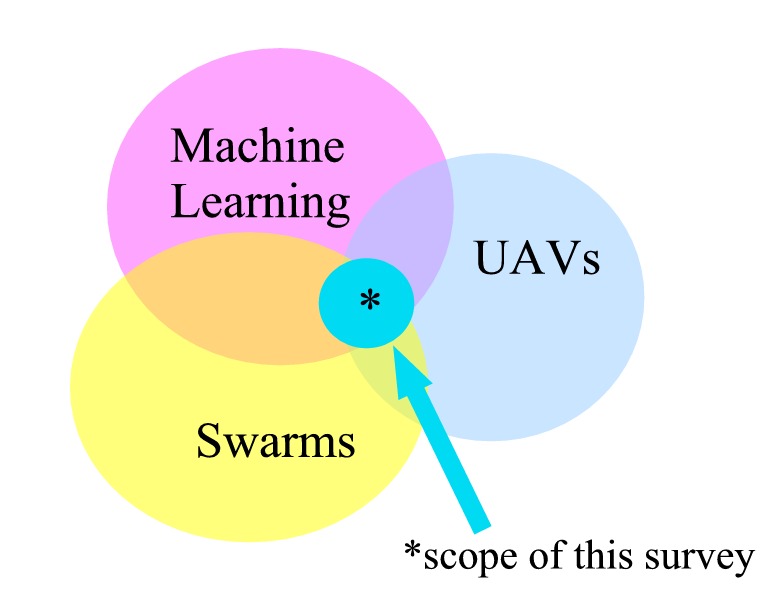}
\end{center}
\caption{Scope of our survey}
\label{figure:survey}
\end{figure}

The contribution and novelty of our survey compared to the surveys mentioned previously and summarized in Table~\ref{table:survey} are based on the fact that we concentrate on using ML methods for UAV flock management: flock formation, flock management, and resource and task allocation in UAV flocks. Thus,
 our scope is in the intersection among ML, swarm intelligence (SI), and studies relating to UAVs, which none of the previous surveys have encompassed. Figure \ref{figure:survey} illustrates this point.

\section{Flocks Formation and Maintenance}
\label{section:formation}

The possible co-existence of large swarms of aerial vehicles has logically led to the necessity of grouping them into flocks. The grouping of UAVs into flocks may be useful for several applications, such as UAV-based communication networks, wireless sensor networks (WSNs), information gathering, image collection, and military applications. The clustering of UAVs implies many tasks, such as dividing them into clusters, where each cluster has a cluster head (CH) that is responsible for data aggregation and transmission to the BS. 

The first problem that arises when dealing with UAV organization into flocks is that of flock formation; that is, how to cluster the UAVs into flocks. This includes (a) the affiliation of each UAV to a certain flock; (b) the internal topology within the flock; (c) the internal routing protocol used in the flock; (d) the required updates of the internal and external structure of the flock given external changes; (e) task allocation within the flock; and (f) task allocation updates given external changes.

The aim of this section is to survey ML methods for UAV flocking. Clustering methods based on bio-inspired algorithms and particle swarm optimization (PSO) are presented in Section~\ref{subsection:formation:swarms}, whereas the issue of how ML methods can be applied to clustering formation is investigated in Section~\ref{subsection:formation:ML}.

\subsection{Bio-inspired Clustering Algorithms}
\label{subsection:formation:swarms}

Based on the fact that flocking behavior occurs widely in nature, several studies have suggested the use of bio-inspired algorithms to achieve efficient flocking behavior. Owing to their popularity, we describe these in a separate section.

The PSO algorithm is an evolutionary computing
technique that is modeled according to the social behavior of a flock of
birds. In the context of PSO, a swarm refers to numerous
potential solutions to the optimization problem, where each
potential solution is known as a particle. The aim of the
PSO is to determine the particle position that results in the best
evaluation of a given fitness function. Regarding the clustering challenge, Latiff et al.
\cite{Latiff} were the first to present energy-aware clustering for WSNs using a PSO algorithm. They
defined a cost function that considers the intra-cluster distance, available energy of each node, and each CH candidate. The simulation results
indicated that the proposed protocol using the PSO algorithm provided
a higher network lifetime, delivered more data, and produced better clustering stations compared to LEACH and LEACH-C.

Following Latiff et al., several variations of the basic PSO algorithm were suggested.
Kuila and Jana \cite{Kuila} presented linear programming/nonlinear programming (LP/NLP) formulations of the clustering and routing problems in WSNs, followed by two proposed algorithms
based on PSO. The clustering algorithm considered the energy conservation of the nodes through load balancing, whereas the routing algorithm considered a tradeoff between the transmission distance and number of hop counts. In the clustering phase, the routing overheads of the CHs were considered for balancing their energy consumption.

Suganthi and Rajagopalan \cite{Suganthi} considered the problem of the construction of energy-effective multi-swarms in dynamic environments and presented variations of PSO that are specifically formulated for excellent functioning in dynamic settings. The primary notion is the extension of single-population PSO as well as charged PSO techniques through the construction of interactive multi-swarms. They also proposed updating as well as recalculating algorithms for connected dominating sets given changes in the ad-hoc wireless network topology.

Collotta et al. \cite{Collotta} 
proposed a fuzzy logic-based mechanism for wireless sensor networks, which defines the sleeping time of the sensor devices. A PSO algorithm was introduced to obtain the optimal values and parameters of the proposed fuzzy logic controller so as to achieve the best results regarding the battery life of the sensor nodes.


A hybrid approach for the clustering scheme was proposed by Aftab et al. \cite{AftabZhang} for drone-based
cognitive IoT, which uses a hybrid mechanism of glowworm swarm optimization \cite{WangGlowworm} and the dragonfly algorithm \cite{Mirjalili}.
Whereas cluster formation and CH selection
were performed using glowworm swarm optimization, the swarm joining and cluster management mechanisms were inspired by the dragonfly algorithm. Tracking methodology was also proposed using the dragonfly algorithm
rules, which aided in more effective management of the cluster topology. The cluster maintenance was performed by a mechanism to identify dead cluster members. Finally, an efficient route selection function was proposed.

Arafat and Moh \cite{ArafatClustering} suggested clustering and CH selection methods based on PSO for reducing the energy consumption and balancing the energy utilization, thereby increasing the network lifetime.
According to the PSO method, after computing the search region, a particle is assigned to a subswarm, which is known as a local swarm. The local swarm overlapping, overcrowding, and convergence are verified before the next iteration starts.
After obtaining the node location information, the node is assigned into a cluster based on the nearest location point.
In the second phase, the cluster sizes are balanced by moving the UAVs from an excessively large cluster to their nearest cluster. The fitness values for the CH selection in the PSO algorithm are based on the weighted sum of three parameters: the intra-cluster distance, UAV energy level, and inter-cluster distance.

Ganesan et al. \cite{Ganesan} aimed to provide an effective solution to elect the CH among multiple drones at different periods based on the various physical constraints of the drones. The elected CH acts as the decision maker and assigns tasks to the other drones. In the case where the CH fails, the next eligible drone is re-elected as the leader.
The CH is elected dynamically based on the following three parameters: its current position (closer to the BS and all other drones), residual energy, and velocity, using a method based on PSO. This elected leader CH alone communicates with the BS and other drones, thereby decreasing the communication energy, which in turn extends the network lifetime.
Clusters are formed using spider monkey optimization based on the proximity of the drones to one another, their connectivity to other nodes (using RSSI), and the residual energy of the drones. The clusters that are formed will have equal average residual energy for increasing the network lifetime further.
In the above study, the simulation demonstrated that the results of the proposed distributed method were more efficient than the results of the PSO-C algorithm, in which the fitness value is based only on the residual energy, in terms of extending 
the network lifetime by reducing the communication energy consumption.

Azoulay and Reches \cite{AzoulayReches,AzoulayRechesJournal} developed a multi-agent (MA) blackboard-based protocol to form clusters of self-interested UAVs with close routes. According to the proposed protocol, the formed clusters are saved in a centralized blackboard, and any new UAV can either join an existing flock or create a new one and publish it in the centralized blackboard. The simulation results of this approach demonstrated a load reduction in crowded environments.

Another bio-inspired clustering algorithm for highly dynamic large-scale ad-hoc UAV networks was suggested by Yu et al. \cite{Yu}, which transplants the foraging model of physarum polycephalum to the field of highly dynamic large-scale ad-hoc UAV networks to adapt to these networks, and conducts cluster formation and maintenance effectively. They demonstrated through simulations that their algorithm outperformed the classical clustering algorithm in terms of the average link connection lifetime and average CH lifetime, which could make the cluster structure more stable.

\subsection{ML Methods for Flock Formation}
\label{subsection:formation:ML}

As highlighted in the literature, the problem of determining optimal clusters, even in static environments, is known to be NP-hard \cite{AgarwalNPH,LuoZhang}; thus, the different methods suggested for relatively static or highly dynamic networks are heuristics. Given that ML exhibits good performance in various NP-hard problems \cite{KhalilNPH}, several studies have suggested the use of advanced ML methods to handle the clustering challenge.
The basic clustering problem is a variation of the dominating set problem \cite{Butenko}. A dominated set of graph G is a subset S of the vertex set of G, in which each vertex in G is either in S or adjacent to a vertex in S. A minimum dominating set is a dominating set of the smallest size in a given graph. The size of the minimum dominating set is known as the domination number of the graph. In wireless sensor applications, each sensor is considered as a node and an edge exists between two nodes if and only if they have a direct connection to one another. Moreover, the set of the CHs is the dominated set of the WSN graph. 
He et al. \cite{HeDS} suggested the use of a Hopfield neural network to minimize the weakly connected dominating set for the self-configuration of WSNs.
They explored
the convergence procedure by simulating the subsequent
model and tested the model on complete graphs to observe
the robustness of the parameters in the model. Furthermore, they presented an improved hill climbing algorithm
with directed convergence, heuristic adjustment, and threshold compensation. They demonstrated the efficiency of their method by means of simulation and investigated the effect
of the transmission radius on the cluster sizes. The results
indicated that it is important to select a suitable transmission radius to ensure that the network has good stability and a long lifespan.

When viewing the clustering and head selection problems along the timeline, the decision in one period may influence the future environmental state and future decision making; thus, RL can be applied to solve the clustering problem over time.
Pan et al. \cite{PanKMeans} proposed an energy-aware CH selection method based on binary PSO and K-means to extend the network lifetime. The selection criteria of the objective cost function were based on minimizing the intra-cluster distance as well as the distance between the CHs and BS, and optimizing the energy consumption of the entire network. Moreover, the sensor nodes were divided into several clusters based on the K-means algorithm at the beginning, which could reduce the complexity of the overall algorithm. They demonstrated by simulation that their technique could outperform LEACH-C and PSO-C in terms of the network lifetime.

Yuan et al. \cite{Yuan17} proposed a genetic algorithm (GA)-based network clustering method that provided a framework for dynamically optimizing wireless sensor clusters. The residual energy, expected energy expenditure, distance to the BS, and number of nodes in the vicinity were used as the arguments for the fitness function, where each GA chromosome represented a designation map of CHs. Given the CHs, the cluster members were formed following the nearest neighbors rule. In each transmission round, the network structure was updated dynamically to achieve a balance of the residual energy of the sensor nodes, and hence, to extend the network longevity.

Tolba and Alarifi \cite{TolbaRL} proposed an RL-based technique known as adaptive Q-learning (AQL) for clustering in a distributed sensor network to improve the network performance with a minimum energy–overhead tradeoff.
AQL operates in two distinct phases, namely those for CH selection and forwarder selection.
The decision-making system is used to qualify nodes based on their past behavior over transmission. AQL improves both the inter- and intra-cluster communication.
The experimental results demonstrated the effectiveness of the proposed learning technique by improving the network lifetime with a high request–response rate, thereby
minimizing the delay, overhead, and request failures.

Finally, Govindara and Deepa \cite{Govindaraj} considered a WSN for the IoT and proposed a capsule neural network architectural model to achieve better performance by minimizing the network energy overhead for the WSN-based IoT. The clustering process in sensor networks contributes to improving the network quality by controlling the energy consumption rate and improving the data accuracy rate. The above authors aimed to develop a new capsule neural network model that can maintain the network energy at an optimal level, leading to improved throughput and higher accuracy with a reasonable network overhead. The capsule neural network architecture is a plausible neural model and has been proven to be effective for routing and optimization operations wherein the activation of the capsule is calculated at the forward pass time.

The idea is to add structures known as “capsules” to a convolutional neural network (CNN) and to reuse output from several of those capsules to form more stable representations (with respect to various perturbations) for higher capsules.
The simulation results that were obtained established the reliability and effectiveness of the proposed capsule neural network learning for the energy optimization of the IoT in sensor networks compared to existing methods in the literature.

To clarify the various algorithms used for clustering, Table ~\ref{table:ML:Clustering} summarizes the
studies that have used bio-inspired and ML methods to solve the clustering challenge.

\begin{table}
\caption{Creation and maintenance of flocks and clusters (C = centralized, D = distributed)}
\label{table:ML:Clustering}
\begin{center}
  \begin{tabular}{ | l | l | l | l | l |}\hline
 Publication & Challenge & Optimization & ML & Approach  \\ 
   &   & &  Method & \\ \hline 
Latiff et al.   & Clustering & intra-cluster distance,   & PSO & D \\ 
 \cite{Latiff} &  &  Energy stability&   &  \\ \hline
Kuila and Jana  & Clustering,  & energy conservation, and  & LP/NLP,  & D \\
 \cite{Kuila} &    Routing &  load balancing &  PSO & D \\ \hline
 Suganthi and   & Multi-swarm
  & energy effectiveness & PSO & D \\ 
   Rajagopalan \cite{Suganthi} &  Construction &    &   & \\ \hline
   Collotta et al. \cite{Collotta} & WSN clustering   & battery life & Fuzzy logic PSO& D \\ \hline
   Aftab et al.  & Drone cluster   & clustering construction time  & Glowworm swarm  & D\\ 
    \cite{AftabZhang} & Management  &  and lifetime,  &  optimization & \\ 
      & and routing  &  energy consumption &   & \\ \hline
Arafat and Moh \cite{ArafatClustering}& Clustering & network lifetime & PSO & D\\ \hline
Ganesan et al. & CH election, & link lifetime, & PSO+MSO & D\\ 
\cite{Ganesan}&  Clustering&   cluster lifetime &  & \\ \hline
Yu et al.  & UAV  & Connection lifetime,  & foraging model of      & D\\ 
  \cite{Yu}& clustering &   CH lifetime & Physarum   &  \\ 
  &  &    &   polycephalum &  \\ \hline \hline
He et al. \cite{HeDS}&  WSN clustering  & stability, lifetime& Hopfield NN & C\\ \hline 
Pan et al. \cite{PanKMeans}& Clustering & energy consumption & PSO+K-means & C/D\\ \hline 
Yuan et al. \cite{Yuan17}& Clustering & network longevity & GA & C\\ \hline 
Tolba and Alarifi \cite{TolbaRL}& WSN clustering & network lifetime & AQL& C \\ \hline 
 Govindaraj and Deepa & WSN clustering & optimal energy   & CNN   & C\\ 
 \cite{Govindaraj}&   & Maintenance   &  & \\ \hline 
\end{tabular}
\end{center}
\end{table}

\subsection{Flocking Strategies and UAV Coordination}

Several challenges in flock management involve the physical location of the flocks. This includes determining the optimal deployment of UAVs, physical coordination of UAVs, and trajectory optimization. In the following section, we survey several recent studies that have suggested the use of ML methods to address the location-related challenges. We focus on certain geographical challenges of flock formation, which are mainly determined by their localization and mobility, to avoid collisions and to achieve energy efficiency. We consider the following related problems: flocking strategies to avoid collisions between flock members, deployment challenges in UAV-based wireless networks, and multi-UAV path planning challenges. Each related problem is discussed in a separate section and relevant ML methods are surveyed, with each of these challenges discussed and solved.

Given a flying flock of objects, it is important for the flock members to coordinate to avoid cohesion.
The basic flocking challenges of any flock of flying objects working independently were defined as follows by Reynolds \cite{Reynolds}: the birds attempt to stick together, and to avoid collisions with one another and with other objects in their environment. There are three basic rules for maintaining the flocking behavior, as follows: separation: avoid collisions with nearby flockmates; alignment: attempt to match velocity with nearby flockmates; and cohesion: attempt to stay close to nearby flockmates.
Similar to the swarming behaviors that are observed in animals and insects, autonomous coordination among UAVs should be guaranteed. In this section, we consider several studies that have used ML methods to maintain stable flock behavior.

A distributed approach for the flocking problem was suggested by Olfati-Saber \cite{Olfati-Saber} in the area of MA dynamic systems, where the agents were autonomous and no centralized manager existed. Oltafi-Saber considered this problem and suggested several distributed flocking algorithms for this purpose. Two cases of flocking in free space and the presence of multiple obstacles were considered. Moreover, three flocking algorithms were proposed: two for free flocking and one for constrained flocking. A systematic method was provided for the construction of cost functions (or collective potentials) for the flocking, in which the collective potentials penalized deviation from a class of lattice-shaped objects. Several simulation results were presented to demonstrate the performance of the 2D and 3D flocking, split/rejoin maneuver, and squeezing maneuver for hundreds of agents using the proposed algorithms.

Maza et al. \cite{Maza,MazaAWARE} considered civil domains in which multiple aerial robots with sensing and actuation
capabilities were available and presented the AWARE project for the autonomous coordination and cooperation of UAVs.
The different components of the AWARE platform and
scenario in which the multi-UAV missions were carried out were described, including surveillance with multiple UAVs, sensor deployment, and fire threat
confirmation. Key issues in multi-UAV systems, such as distributed task allocation, conflict resolution, and plan refinement, were solved in the execution of the missions.
Constraint-based temporal planning and scheduling were used for solving the planning problems, and the contract net protocol was used to manage the distributed task allocation process among the different UAVs.
They demonstrated by means of experiments with real UAVs that the
developed architecture allowed a broad spectrum of missions to be covered: surveillance, sensor deployment, fire confirmation, and extinguishing.

Quintero \cite{Quintero} developed a novel flocking algorithm that enabled multiple UAVs to locate themselves in the flock to  distribute a given sensing task among the group members, assuming a leader–follower network topology. They focused on the control policy of the followers, and developed a cost function that is a function of the distance and heading with respect to the leader, as well as a stochastic kinematic model that facilitates flocking. Thereafter, they used dynamic programming to minimize the expected cost of each follower. Quintero et al. assumed that there is a predefined flock leader that is known to the entire flock, and their aim was to optimize the distance and heading of the followers with regard to the leader.

Xu et al. \cite{XuDistributed} proposed 
a distributed neuro-dynamic flocking design for ensuring that the UAVs follow three heuristic flocking rules, namely cohesion, separation, and alignment, in an optimal manner, using the neuro-dynamic programming (NDP) technique, First, an innovative cost function was developed by combining the system cohesion, separation, and alignment performance. Subsequently, a novel neural network was proposed to approximate the minimized cost function value by using the Hamilton–Jacobi–Bellman  equation in an online and forward-in-time manner. Thereafter, near-optimal flocking could be achieved by minimizing the estimated cost function.

Hung and Givigi \cite{Hung} proposed the use of a model-free RL method to enhance the autonomous coordination among UAVs in a swarm. They used a leader–follower policy, whereby Peng's $Q(\lambda)$ \cite{PengQ} with a variable learning rate was employed by the followers to learn a control policy that facilitates flocking. The problem was structured as a Markov decision process (MDP), in which the agents were modeled as UAVs that experience stochasticity owing to disturbances such as winds and control noises, as well as weight and balance issues. The learned policies were compared to dynamic programming methods that were solved using stochastic optimal control. The simulation results demonstrated the feasibility of the proposed learning approach for enabling agents to learn how to flock in a leader–follower topology while operating in a non-stationary stochastic environment.

Tsai \cite{Tsai} focused on vision-based collision avoidance for UAVs. First, images from the UAV cameras were fused based on deep CNNs. Thereafter, a recurrent neural network (RNN) was constructed to obtain high-level image features for object tracking and to extract low-level image features for noise reduction. The system distributed the calculation among the multiple UAVs to perform object detection, tracking, and collision avoidance efficiently. To achieve high overall system performance, the study adopted the half total error rate (HTER) accuracy measurement, which is based on the false rejection rate and false acceptance rate.

Jafrai et al. \cite{Jafari} 
proposed a flocking design framework for MA systems using an RL technique, which was appropriate for real-time implementation. The controller that they developed was based on a bio-inspired RL controller, relying on a computational model of emotional learning in the mammalian limbic system. They used this model for practical MA systems in the presence of uncertainty and dynamics.

In the work of Venturini et al. \cite{Venturini}, the authors considered a general MARL framework for the
initial exploration and surveillance of a swarm of independent UAVs. Their scheme followed the framework in which observations of other agents are used to make decisions and to avoid collision, thereby encouraging cooperation. They
defined a deep Q network algorithm. They subsequently demonstrated its
efficiency with limited training, compared it to the look-ahead search heuristic, and showed that the MARL scheme can explore
the environment and reach the targets more rapidly. They also performed a
transfer learning experiment, demonstrating that agents trained on a different map can learn to adapt to a completely new scenario much
faster than when restarting the training from scratch.

Anicho et al. \cite{anicho}
compared the performance of RL and SI methods for solving the problem of coordinating multiple
high-altitude platform stations (HAPSs) for communication
area coverage. Swarm coordination techniques are essential
for developing autonomous capabilities for the control and management of multiple HAPSs. The
authors observed that the RL approach exhibited superior overall peak
user coverage with unpredictable coverage dips, whereas the SI-based approach exhibited lower coverage peaks but better
coverage stability and faster convergence rates.

Sharma and Ghose \cite{Sharma2} developed several basic swarming laws for UAVs.
They demonstrated that when the cohesion rule is applied, an equilibrium condition is reached, in which all of the UAVs settle at the same altitude on a circle with a constant radius.
They proposed a decentralized autonomous decision-making approach that could achieve collision avoidance, and developed algorithms with the aid of these swarming laws for two types of collision avoidance, namely group-wise and individual, in the 2D plane and 3D space.
Their experimental results demonstrated the effectiveness of their approach, in which self-organized flight cluster collision avoidance was successfully achieved by the UAV swarms.

Wu et al. \cite{WuLi} studied the problem of path conflicts for UAV clusters and established a method for calculating the collision probabilities of UAVs under the constraints of the mission space and number of UAVs. In the cluster flight mode, the automatic tracking and prediction of UAV cluster tracks were implemented to avoid path conflicts in the clusters. To address the
inconsistency problem owing to noise, a state estimation method based on the Kalman algorithm was proposed.
The cluster state prediction and collision probability were calculated to prevent the clusters of formation UAVs conflicting on paths during flight. Finally, the simulation results verified the
validity and effectiveness of the proposed method in multi-UAV formation flight planning.


Table
~\ref{table:ML:flocking} summarizes the studies relating to flocking and collision avoidance among UAVs using ML methods.
\begin{table}
\caption{ML applied to flocking and coordination}
\label{table:ML:flocking}
\begin{center}
  \begin{tabular}{ | l | l | l | l | }\hline
 Publication & Challenge & Optimized metric & ML   \\ 
   &   & &  Method \\ \hline 
Olfati-Saber \cite{Olfati-Saber} & Flocking challenges & Connectivity, energy & Distributed flocking  \\ 
  &   &  &  algorithms\\ \hline
Maza et al. \cite{Maza,MazaAWARE} & Architecture & execution cost & Temporal planning,   \\
 &   &  & contract net \\ \hline 
Quintero \cite{Quintero} & Localization & distance and heading & DP\\ \hline
Xu et al. \cite{XuDistributed}  & Ensuring flocking rules & cohesion, separation,   & NDP\\
  &  &  and alignment &  \\ \hline
Hung and Givigi \cite{Hung}  & Coordination & flocking cost function& Q-learning \\ \hline
Tsai \cite{Tsai} & Vision-based collision    & HTER & RNN \\ 
& avoidance  &   &   \\  \hline
Jafrai et al. \cite{Jafari}  & Flocking design & multi-objective properties  & Bio-inspired RL \\ \hline
Venturini et al. \cite{Venturini} & Exploration and  &  target reaching efficiency & Deep Q-learning\\
 &  surveillance&    &  \\ \hline
Anicho et al. \cite{anicho} & Coordinating & coverage & RL and SI\\ \hline
Sharma and Ghose \cite{Sharma2} & Collision avoidance & Swarm size and stability & Swarm laws \\
 & &  & Decentralized alg.\\ \hline
Wu et al.   & Object tracking & noise, & Kalman  \\ 
  \cite{WuLi} &  &  Collision prediction &    algorithm\\ \hline
\end{tabular}
\end{center}
\end{table}

\subsection{UAV Deployment}
\label{section:deployment}
A special case of task allocation involves the positioning of UAVs in wireless communication networks, where the UAVs function as BSs, and their deployment affects the network efficiency and coverage.
In fact, the problem of determining the optimal placement of ABSs
to maximize the coverage is known to be a NP-hard problem \cite{Lyu2017},
and in this section, we survey several ML-based methods for addressing this challenge.

Park et al. \cite{ParkLee}
considered the problem of the optimal deployment of multi-UAVs to provide high throughput for
users with different requirements in the UAV–BS environment.
Based on the air-to-ground path loss model,
the virtual communication environment was established, in which airtime
fairness was applied for equitable time distribution of the channel usage
according to user requirements.
RL was applied to determine the best UAV positions, and a collaborative 
algorithm with modified K-means was employed to distribute users to each
UAV to solve communication overload problems.

In the study of Klaine et al. \cite{Klaine}, the aim was to maximize the number of users covered by the system in an emergency scenario, where the drones were limited by both backhaul and RAN constraints.
For this scenario, Klaine et al.
proposed the use of RL to determine the optimal position of the UAVs.
The proposed solution
was compared to different positioning strategies, such as
deploying the drones in fixed random positions, fixing the drones
around a circle centered in the middle of the area at evenly
spread angles, or deploying the drones in the locations of
hotspots of a previously destroyed network. The results
demonstrated that the Q-learning solution outperformed all other methods
in all considered metrics.
Liu et al. \cite{Liu} proposed a deep RL (DRL) variation known as the deep deterministic policy gradient method for energy-efficient UAV control in the context of providing communication coverage for ground users. The control policy considers the UAV movements in each time slot and the aim is to optimize the communication coverage, fairness, energy consumption, and connectivity.

The deployment decision becomes more complicated when the environment itself, 
the coverage map in particular, is initially unknown, and the effect of each UAV position on the coverage is not known in advance. Zhang et al. \cite{ZhangDeployment} suggested an ML framework based on a Gaussian mixture model and a weighted expectation maximization algorithm to predict the potential network congestion. Based on the  
predicted congestion and traffic, the optimal deployment of the UAVs could be achieved in a manner that minimized the transmit power required to satisfy the communication demand of users in the downlink, while minimizing the power required for the UAV mobility.
Qiu et al. \cite{QiuLyu2020} considered 
the NP-hard problem of maximizing the coverage rate of
N ground users by the simultaneous
placement of multiple ABSs with a limited coverage range.
In their study, Qiu et al. applied the DRL method for representing the state by a coverage bitmap to capture the spatial correlation
between ground users and ABSs, and for effectively learning
the action and reward function given dynamic interactions with the complicated propagation environment.

In the study of Liu et al. \cite{LiuLiuClustering}, the maximization criteria consisted of the ground user opinions.
They formulated the problem of the joint non-convex 3D deployment and dynamic
movement of the UAVs, where the goal was to maximize the sum of
the mean opinion score (MOS) of the ground users. They demonstrated that the problem was NP-hard and proposed 
a Q-learning-based solution to handle the problem.
An algorithm based on a combination of the GA and K-means was used to obtain
the cell partition of the users. Subsequently, a Q-learning-based deployment algorithm was proposed to achieve 3D
placement of the UAVs when the users were static. Finally, a Q-learning-based
movement algorithm was presented to obtain the 3D dynamic movement of the UAVs.
Several recent studies have suggested different types of learning to accelerate the deployment process. Liu et al. \cite{LiuWangFast} developed a fast positioning algorithm for the deployment of UAVs serving as BSs, which achieved the goal of maximizing the sum of the downlink rates in the multi-UAV communication network.
They designed a geographical position information (GPI) learning algorithm to learn the GPI relationship between the users and UAVs and demonstrated that the GPI learning method enabled rapid suboptimal deployment in the case of multiple UAVs and users.

Another aspect that can be considered in optimal UAV deployment is the expectation of user requests.

Liu et al. 
\cite{LiuNavigation} considered the problem of navigation control for a group of UAVs that served as mobile BSs. The UAVs were supposed to fly around a target area to provide long-term communication coverage for ground mobile users. Liu et al. designed a decentralized DRL-based framework to control each UAV in a distributed manner. Their goals were to maximize the temporal average coverage score achieved by all UAVs in a task, maximize the geographical fairness of all considered points of interest (PoIs), and minimize the total energy consumption, while maintaining the UAVs connected and located inside the area borders. They designed the state, observation, action space, and reward in an explicit manner and modeled each UAV using deep neural networks (DNNs).

In the remainder of this section, we consider studies that aimed to deal with a joint optimization problem, namely the deployment challenge that considers user association, power control, and other optimization problems that are involved in implementing a UAV-based wireless network.

Dai et al. \cite{Dai}
investigated the problem of the efficient deployment of UAVs while guaranteeing the quality-of-service requirements.
The UAV played the role of a coordinator to provide a high-quality communication service for
ground users as well as to maximize the benefits of caching. They proposed an RL-based approach to solve the multi-objective deployment problem, while maintaining an optimal tradeoff between
the power consumption and backhaul saving.
Owing to the interdependent relationships among
different UAVs, they adopted RL based on the local
search approach to determine the 3D placement, minimum
transmit power, and cache strategy of each UAV. Finally, the minimum number of UAVs 
was provided together with the efficient deployment scheme.

In another study, Dai et al. \cite{DaiDeploy} leveraged game theory to model the problem of imperfect channel state information and proposed a robust and distributed learning algorithm to identify the multi-UAV plan flight paths while simultaneously gathering local information until the optimal deployment location was determined. They demonstrated via simulation that the suggested distributed learning algorithm converged to a stochastic stable state, which maximized the optimization objective; that is, the sum of the alpha-fairness of all ground terminals.

Chen et al. \cite{ChenCachingSky} studied the problem of the proactive deployment of cache-enabled UAVs for optimizing the quality-of-experience (QoE) of wireless devices in a cloud RAN. In the considered model, the network could leverage human-centric information such as the visited locations, requested contents, gender, job, and device type of users to predict the
content request distribution and mobility pattern of each user. Subsequently, given these behavior predictions,
the proposed approach sought to determine the user–UAV associations, optimal UAV locations, and
contents to cache at the UAVs. The problem was formulated as an optimization problem, the goal of which was
to maximize the QoE of the users while minimizing the transmit power used by the UAVs. To solve this
problem, they proposed the use of conceptor-based echo state networks (ESNs). The ESNs
could effectively predict the content
request distribution and mobility pattern of each user when limited information on the states of the users and
network was available. Based on the predictions of the content request distributions and mobility
patterns of the users, the user–UAV association, UAV locations, and
contents to cache at the UAVs could be optimally determined.

Several studies considering the deployment challenge have suggested the use of clustering algorithms
to cluster the ground user, and subsequently to determine the UAV deployment based on the clustered users. Several studies proposed running this scheme iteratively until convergence is reached.
Kang et al. \cite{Kang}  proposed a
cluster-based UAV deployment scheme to reduce the data
traffic as well as service delays and to improve the coverage of BSs. User groups were formed using the K-means clustering algorithm.
Thereafter, the optimal UAV locations were determined given the user clusters.
Furthermore, a long short-term memory (LSTM)-based caching scheme was proposed to cache the popular contents on UAVs. 
Yang et al. \cite{YangPan} considered the sum power minimization problem via jointly optimizing the user association, power control, computation capacity allocation, and location planning in a UAV-based network. They 
proposed a low-complexity algorithm for solving these subproblems iteratively. The compressive sensing-based algorithm was proposed for the user association subproblem. For the computation capacity allocation subproblem, the optimal solution was obtained in a closed form. For the location planning subproblem, the optimal solution was effectively obtained via a 1D search method. Finally, to obtain a feasible solution for this iterative algorithm, a fuzzy C-means clustering-based algorithm was proposed.

Koushik et al. \cite{Koushik} studied an aerial network in which certain UAVs were used as gateway nodes to aggregate the data from other UAVs and send these to a nearby aircraft that acted as a control node for the UAVs in the swarm.
To handle the dynamic swarm topology and time-varying link conditions, they designed the following solution: A deep Q-learning algorithm was used to determine the optimal links between nodes, and an optimization algorithm was applied to fine-tune the position of the UAV node locally to optimize the overall network performance.

Finally, Nguyen et al. \cite{Nguyen} 
considered both the 
deployment and resource allocation challenges for a distributed UAV wireless network in disaster situations. They proposed a rapid user clustering model based on the K-means procedure for the UAV network. Furthermore, they proposed distributed real-time power allocation to maximize the end-to-end sum rate and embedded programming into the UAV devices for rapid recovery of the network to support a large number of users in disaster communication.

It can be concluded that the various studies have considered different aspects of the UAV deployment challenge. They differ in terms of the maximization criteria, data availability, and additional problems that are considered jointly with the deployment decision.
Table~\ref{table:ML:deployment} summarizes the studies dealing with deployment, path planning, and collision avoidance using ML methods.
\begin{table}
\caption{ML application for deployment}
\label{table:ML:deployment}
\begin{center}
  \begin{tabular}{ | l | l | l | l | }\hline
 Publication & Challenge & Optimized metric & ML   \\ 
   &   & &  Method \\ \hline 
Park et al. \cite{ParkLee} & Deployment & throughput & RL, K-means  \\ \hline 
Klaine et al. \cite{Klaine} & Deployment  & coverage & Q-learning         \\  \hline 
Liu et al. \cite{Liu} & UAV movements & coverage, fairness,  & DRL \\
 &  &  energy consumption,  &  \\
 &  &  and connectivity &  \\ \hline 
Zhang et al. \cite{ZhangDeployment} & Congestion prediction & power minimization& Gaussian  model \\ \hline  
Qiu et al. \cite{QiuLyu2020}  &  Action and reward function & coverage rate & DRL \\ \hline 
Liu et al. \cite{LiuLiuClustering}  & Deployment and movement &  mean opinion score& Q-learning, \\ 
 & &  & GA \\ 
& &  & K-means \\  \hline 
Liu et al. \cite{LiuWangFast} & Deployment & sum rate & DNN \\ \hline 
Liu et al. \cite{LiuNavigation} & Navigation & coverage, fairness,
& Decentralized DRL   \\
 &  & energy minimization &   \\ \hline 
Chen et al. \cite{ChenCachingSky}  & User requests and  & QoE & ESN\\
  &  mobility prediction &  & \\ \hline 
Dai et al. \cite{Dai} & Deployment & quality of service & RL + local search \\ \hline 
Dai et al. \cite{DaiDeploy}  & Path planning, deployment & alpha-fairness &  Distributed learning  \\ \hline 
Kang et al. \cite{Kang}  & Deployment, caching,&  reducing delays & Clustering, LSTM \\
  & &  improved coverage &  \\ \hline 
Yang et al. \cite{YangPan}  & User association, power control,   & power minimization &  Iterative algorithm,  \\
   &      &   &   clustering\\
  &   computation capacity allocation,  &   &   \\ 
  &   and location planning  &   &   \\ \hline 
Koushik et al. \cite{Koushik} & Routing   & throughput & Q-learning, \\ 
 & and deployment &  & Optimization  \\  \hline 
Nguyen et al. \cite{Nguyen} & Deployment and  & throughput  & K-means, \\
 & resource allocation &   & Distributed algorithm\\ \hline
\end{tabular}
\end{center}
\end{table}

\subsection{Trajectory Optimization and Navigation}
\label{section:path}

In this section, we extend our survey by reviewing the geographical challenges of flock management. We deal with the trajectory and path planning challenges and review recent studies that have aimed to address these challenges using ML methods.
Trajectory planning refers to moving from point A to point B without collisions, which can be computed using both discrete and continuous methods. Trajectory planning is considered as a major topic in robotics as it paves the way for autonomous vehicles.

Several recent studies \cite{Temizer,Ragi2013,Dabbiru} considered the issue of path planning and collision avoidance in UAV flocks using the MDP.
Bayerlein et al. \cite{Bayerlein} 
leveraged the use of RL, where the UAV
acted as an autonomous agent in the environment, to learn the
trajectory that maximized the sum rate of the transmission during
the flying time.
Movement decisions were made directly by a neural
network. The algorithm did not require explicit information regarding
the environment and could learn the network topology to improve the system-wide performance.

Zeng et al. \cite{ZengNavigation} formulated the UAV trajectory optimization problem as the minimization of the weighted sum of its mission completion time and the expected communication outage duration. They demonstrated that the formulated problem could be transformed into an equivalent MDP. Thereafter, it was proposed to use an RL technique, such as the classical Q-learning approach, to learn the UAV action policy, namely the flying direction in this context.
As the output of the MDP involves a continuous state space that essentially has an infinite number
of state–action pairs, they applied a DNN to approximate the Q function. They used the dueling network architecture with multi-step learning to train the DNN. The proposed DRL-based trajectory design did not require any prior knowledge regarding the channel model or propagation environment. It only utilized the raw signal measurement at each UAV as the input to improve its radio environmental awareness.

In the study of Hu et al. \cite{HuRLTrajectory}, a decentralized RL method was adopted to solve the
UAV trajectory design problem. To
coordinate multiple UAVs performing real-time sensing tasks,
they first proposed a sense-and-send protocol, and analyzed the
probability of successful valid data transmission using nested
Markov chains. Thereafter, they formulated the decentralized trajectory
design problem and propose an enhanced multi-UAV Q-learning
algorithm to solve this problem. The simulation results demonstrated that the
proposed enhanced multi-UAV Q-learning algorithm converged
faster and achieved higher utilities for UAVs in real-time
task-sensing scenarios.

Multi-ant colony optimization (ACO) was suggested by Cekmez et al. \cite{Cekmez} for the obstacle avoidance path planning challenge. 
According to the multi-ACO method, numerous ant colonies attempt to determine an optimal solution cooperatively by exchanging their valuable information with one another. Cekmez et al. aimed to implement obstacle avoidance UAV path planning using this algorithm, and they experimentally demonstrated that this approach achieved effective path planning for UAVs compared to a single-colony ACO approach.

\subsubsection{Trajectory Design with Additional Challenges}

In recent studies, the UAV trajectory challenge in wireless communication networks has also been discussed in combination with the related resource allocation challenges.

Challita et al. \cite{ChallitaDRL,ChallitaInterference1} considered a UAV cellular network and aimed to minimize the interference caused by
the ground network, as well as the wireless transmission
latency. They modeled the problem as a distributed multi-UAV game, in which the objective of each UAV was to learn its path, transmit power level, and association vector autonomously and jointly. For the proposed game, the cell association vector,
trajectory optimization, and transmit power level of the UAVs were closely coupled with one another and their
optimal values varied according to the network dynamics.
Challita et al. suggested the use of DRL based on ESN cells for optimizing the trajectories of multiple cellular connected UAVs in an online manner to achieve convergence to equilibrium.
The deep ESN architecture was trained to allow each UAV to map each observation of the network state to an action, with the aim of minimizing a sequence of time-dependent utility
functions. Each UAV used the ESN to learn its optimal path, transmit power level, and cell association
vector at different locations along its path. The proposed algorithm was demonstrated to reach subgame perfect
Nash equilibrium (SPNE) upon convergence.

Liu et al. \cite{LiuLiu} addressed the challenge of the joint trajectory design and power
control of UAVs that served as BSs. Their goal was to improve the user throughput, while satisfying the user rate requirement. Their solution was based on an MA Q-learning based placement algorithm to determine the initial deployment of the UAVs. Furthermore, an ESN-based prediction algorithm was used to predict the mobility of users and an MA Q-learning-based method was considered for the trajectory acquisition and power control algorithm for the UAVs.

The objective in the study of Khamidehi et al. \cite{Khamidehi} 
was to determine the trajectory of multiple ABSs, such that the sum rate of the users served by each ABS was maximized. For this purpose, along
with the optimal trajectory design, they also considered the optimal power and subchannel allocation, which are of great importance for supporting users with the highest possible data rates. Thus, they divided the problem into two subproblems:
ABS trajectory optimization, and joint power
and subchannel assignment. Subsequently, they developed a distributed Q-learning-based algorithm to solve these subproblems efficiently, which did not
require a significant amount of information exchange between
the ABSs and core network. The simulation results demonstrated
that although Q-learning is a model-free RL technique, it has a remarkable capability to train
 ABSs to optimize their trajectories based on the
received reward signals, which carry information
from the network topology.

Qie et al. \cite{QieShi} 
\cite{Qie} considered a multi-UAV target assignment scenario, where a flock of UAVs was supposed to fly to targets that were distributed at different
locations with the shortest total flight distance, with
certain fixed threat areas that the UAVs could not enter. Of course, collision
avoidance between UAVs was required, and it was assumed that there was only one
type of target and all UAVs were identical. This combinatorial optimization problem included two main subproblems: target assignment and path
planning. They proposed a simultaneous target assignment and path planning (STAPP)
method based on an MA deep deterministic policy gradient (MADDPG) algorithm, which is a type of MARL algorithm. In STAPP, the multi-UAV target assignment problem was first constructed as an MA system. Thereafter, the MADDPG framework was used to train the system to solve the target assignment and path planning simultaneously, according to a corresponding reward structure. The proposed system could deal with dynamic environments effectively, as its execution required only the locations of the UAVs, targets, and threat areas.

Table
~\ref{table:ML:trajectory} summarizes the part of our survey related to the trajectory design of UAVs using ML methods.
\begin{table}
\caption{ML application for deployment and positioning }
\label{table:ML:trajectory}
\begin{center}
  \begin{tabular}{ | l | l | l | l | }\hline
 Publication & Challenge & Optimization & ML   \\ 
   &   & Criteria &  Method \\ \hline 
Bayerlein et al. \cite{Bayerlein}& Trajectory &sum rate & Q-learning   \\ \hline 
Zeng et al. \cite{ZengNavigation} &  Trajectory   & Weighted sum of  & RL, DNN   \\ 
 & optimization    &  mission completion &     \\  \hline 
Hu et al. \cite{HuRLTrajectory}& Trajectory design   & successful Transmissions &  Multi-UAV Q-learning  \\  \hline 
Cekmez et al. \cite{Cekmez} &  Path planning & fitness & Multi-ACO\\   \hline 
Challita et al. \cite{ChallitaInterference1} & Subgame  & routing, power& Distributed DRL  \\ 
&perfect equilibrium &&\\ \hline \hline 
Liu et al. \cite{LiuLiu}& Trajectory design  & Throughput & MA     \\ 
 &  and power control &   &  Q-learning    \\  \hline 
Khamidehi et al. \cite{Khamidehi}&  Trajectory  & sum rate &  Distributed Q-learning   \\ \hline 
Qie et al. \cite{QieShi} & Target assignment   & Shortest flight distance&   MADDPG \\  
&   & &     \\  \hline 
\end{tabular}
\end{center}
\end{table}

\section{ML Methods for Resource Allocation in Aerial Flocks}
\label{section:resource}

A challenging issue when considering flock management is the allocation of communication resources among the flock members, given that the flock members organize an ad-hoc mesh network for their internal and external communication tasks.
Note that D2D network challenges arise in FANETs as well as in other ad-hoc networks, such as VANET, fixed sensor networks, and even any wireless network. Thus, in this section, we survey not only studies that have considered FANETs, but also those that can easily be applied to FANETs. We review these studies for the sake of completeness, as we ought to provide designers of UAV flocks with all of the elements that can be used to develop solutions to their specific scenarios.

Many types of resources can be allocated among flock members, as described in the following:
\begin{itemize}
\item Transmission time: This can be managed via time-division multiple access (TDMA), which allows several users to share the same frequency channel by transmission at different time slots.
\item Frequency channel: This can be managed via frequency-division multiple access (FDMA), which consists of dividing the available bandwidth into several non-overlapping frequency channels, where each channel is allocated to a different user.
\item Power control: This is important to achieve the optimal transmission power for each D2D transmission so as to reduce the interference on other concurrent transmissions, while maintaining a sufficiently strong received signal at the target receiver. It is also important for avoiding the near–far problem (mainly in the CDMA technique) and saving battery life.

\item Multiple input, multiple output (MIMO) antennas: These enable the control of a new dimension, namely the possibility of focusing on a specific target via beam-forming, and also improve the received signal via signal processing. 
\end{itemize}

Three different situations arise when considering resource allocation problems in UAV flocks or other D2D mesh networks:
\begin{itemize}
\item The cluster is completely covered by a ground BS.
\item The cluster is partially covered by a ground BS.
\item The cluster does not detect any ground BS in the entire area.
\end{itemize}
Clearly, the most difficult situations are those in which no BS is detected, as well as those that may occur in uninhabited areas, areas following a disaster, or for flying flocks. In such situations, all of the decisions should be made by the flock members, with no involvement from external management.

Another level of distinction is in the goal function that one is attempting to maximize.
Cui et al. \cite{CuiSpatial} specified four fundamental energy-efficient
metrics that have received the most research attention in the literature:
\begin{itemize}
    \item GEE: the system global energy efficiency;
    \item WSEE: the weighted sum energy efficiency;.
    \item WPEE: the weighted product energy efficiency; and
    \item WMEE: the weighted minimum energy efficiency.
\end{itemize}

Additional optimization criteria include::
\begin{itemize}
    \item maximizing the system throughput;
    \item maximizing the quality of service, in a manner that is compatible with the request types;
    \item minimizing the average latency; and
    \item minimizing the maximal latency of any device.
\end{itemize}

In general, the common resource allocation problems are known to be NP-hard \cite{AgarwalNPH,LuoZhang}.
Thus, over the years, various suboptimal resource allocation algorithms have been developed to deal with such problems \cite{ShiRaz,CynaraDistributed}.

Another difference among resource allocation policies is how decisions are made regarding the distribution of resources, for which three decision-making styles are used:
\begin{itemize}
\item Centralized resource management: One of the UAVs, known as the CH, is responsible for the decision-making process. Centralized management can also be implemented by the ground BS in situations where the cluster is covered or partially covered by such a BS. 
\item Decentralized decision-making process: The decision-making process is handled in a distributed manner by the network/flock members.
Coordination among the decisions of the members is achieved using spectrum sensing and message passing.
\item Hybrid approach: Decisions are made in a manner that combines the distributed decisions of the flock members with the central management of the manager.
\end{itemize}

In the remainder of this section, we describe several studies that have leveraged ML methods to address the challenge of resource allocation in ad-hoc networks. For each considered study, we mention the challenge and the main details of the ML method or methods used to address the challenge. We structure the section according to the ML method used to solve the resource allocation problem.

\subsection{Deep Learning-Based Methods}

Several recent studies in communication resource management have suggested the use of deep learning models to determine efficient optimal solutions to the scheduling and control decisions made in a UAV wireless network. Using deep learning schemes, a multilayer neural network is trained, where the network inputs are the network state (in a given representation), and the output that should be trained is the resource allocation decision. A DNN can be trained either with a supervised training scheme using resource allocation solutions that are calculated using any optimization method, or with an unsupervised scheme by calculating the value of the neural network output and optimizing this value by changing the neural network weights. In the following, we describe several works of both types. 

Sun et al. \cite{SunLearning} proposed a deep learning-based scheme for real-time resource management over interference-limited wireless networks. In particular, they suggested the use of deep learning to approximate the behavior of a certain optimization algorithm. Their theoretical results indicated that it is possible to learn a well-defined optimization algorithm highly effectively by using finite-sized DNNs. To demonstrate their claim, they constructed such a DNN for the power control problems over either the IC or the IMAC channel and trained the DNN to approximate the behavior of the WMMSE algorithm effectively \cite{ShiRaz}.

Cui et al. \cite{CuiSpatial} used a deep-learning approach  for the scheduling of interfering links in a dense wireless network with full frequency reuse. They employed a novel neural network architecture that takes the geographic spatial convolutions of the interfering or interfered with neighboring nodes as the input over multiple feedback stages to learn the optimal solution. To ensure fairness, they proposed a scheduling approach that uses the sum rate optimal scheduling algorithm over subsets of links for maximizing the proportional fairness objective over the network. In their study, Cui et al. suggested two methodologies for the neural network training: the first is a supervised learning process, in which the network is trained by suboptimal solutions to the resource allocation problem, as computed by FPLinQ \cite{ShenYuFPLinQ}, which is an algorithm based on the fractional programming approach, and the second is an unsupervised training process, in which the sum rate is supposed to be maximized. According to their comparison, the unsupervised sum rate scheme outperformed the supervised learning scheme for layouts containing links with similar distances.

Matthiesen et al. \cite{Matthiesen2} developed a deep learning system for energy-efficient power control in wireless networks. They used an optimal reduced-complexity branch-and-bound procedure to determine the globally optimal power policy for a large set of situations, and subsequently used the solved set as the training set for a DNN. The numerical results demonstrated that the proposed method achieved near-optimal performance in suggesting the optimal power control policy.
Azoulay et al. \cite{KirilPowerControl} used a DNN to learn the optimal power control and to request scheduling in a MANET cluster, were the optimization problem was mathematically defined and the DNN was trained by examples that were solved by an optimal solver.
The simulation results revealed that the DNN could serve as a computationally inexpensive component of expensive optimization algorithms in real-time tasks, providing a very good approximation solution for these problems with a very low average run time.

Ahmed et al. \cite{AhmedDeepRL} developed a deep learning-based resource allocation model with the objective of maximizing the total network throughput by performing joint resource allocation (for both the power and channel). They used a supervised learning approach to train the model and obtained the training data by solving the non-convex optimization problem using a GA.
However, despite the similarities between WSNs and UAV networks, certain differences should be considered, such as the high mobility and large distances of the UAVs in an ad-hoc UAV wireless network. Thus, in the remainder of this section, we focus on ML methods that have been used for resource allocation in UAV wireless networks.

Chen et al. \cite{Chen2} considered the problem of joint caching for a network of
cache-enabled UAVs that served wireless ground users over the long-term evolution (LTE) licensed
and unlicensed (LTE-U) bands.
They proposed a distributed algorithm
based on the ML framework of the liquid state machine (LSM) to predict the user content request distribution and to enable the UAVs to
select the optimal resource allocation strategies that maximized the number of users with stable queues depending on the network states autonomously. The optimization problem that was solved was a combined user association, spectrum allocation, and content caching problem.

CNN methods were used by Lee at el. \cite{LeeKim} to learn the efficient transmit power control strategy for maximizing either the spectral efficiency or energy efficiency. Moreover, they proposed a form of deep power control that could be implemented in a distributed manner with local channel state information, allowing the signaling overhead to be greatly reduced. Through simulations, they demonstrated that the deep power control could achieve almost the same or even higher spectral efficiency and energy efficiency compared to a conventional power control scheme, with a much lower computation time.


In summary, when considering the use of deep learning for any resource allocation management, the above related studies have suggested: (a) how to present the input, (b) the value to be optimized, and (c) how to train the network. Given all of the above, the recent studies succeeded in solving different scheduling and resource allocation problems by using deep learning with algorithms based on steps (a) to (c). Previous works have demonstrated the benefit of the deep learning approach in solving complex optimization problems, and in particular, complex resource allocation problems.

\subsection{RL and DRL}

Owing to the high dynamics of UAV networks, several studies have considered RL-based methods to manage the resource allocation in these environments. RL methods can consider the effect of each episodic decision on the future network behavior. This is particularly important when the goal is to maximize an objective function such as the quality of service or latency, where the aim is to avoid long queues and to fulfil the user requirements as soon as possible, because in this situation, each decision may be crucial to avoid long queues in the future system behavior as a result of current decisions.  RL and DRL methods can be applied to enable resource management decisions that consider the system behavior over time.

Zhang et al. \cite{Zhang} proposed an efficient energy and radio resource management framework based on intelligent power cognition for solar-powered UAVs. In particular, they suggested the use of RL for adjusting the energy harvesting, information transmission, and flight trajectory to improve the utilization of solar energy, with two primary goals: staying aloft over a long period and achieving high communication performance.

Hu et al. \cite{Hu} suggested the use of RL for the following three essential parts in the cellular UAV network: protocol design, trajectory control, and resource management. They presented a distributed sense-and-send protocol to coordinate the UAVs, proposed the use of a Q-learning algorithm for the trajectory control and resource management, and introduced different types of RL approaches
and their applications, including the user association, power
management, and subchannel allocation.

Cao and Liang \cite{Cao} proposed a DRL framework for channel and power allocation in a UAV communication system, in which the UAVs were used as BSs. With the proposed framework, the UAV stations could allocate both the channels and transmit power for the uplink transmission of the IoT nodes, with the aim of maximizing the minimum energy efficiency of the IoT nodes.

Ciu et al. \cite{Cui,CuiLiuML}
addressed the challenge of the autonomous resource allocation of multiple UAV communication networks with the goal of maximizing the long-term rewards. To model the uncertainty of the environments, they formulated the resource allocation problem as a stochastic game, in which each UAV becomes a learning agent and each resource allocation solution corresponds to an action taken by the UAVs.
They developed an MARL-based resource allocation algorithm to solve the formulated
stochastic game of multi-UAV networks. Specifically, each UAV, as an independent learning
agent, discovered its best strategy according to its local observations using 
the Q-learning algorithm.
All agents implemented a decision algorithm independently but shared a common structure based on Q-learning.
They provided convergence proof of the proposed MARL-based
resource allocation algorithm, and demonstrated by simulations that the proposed MARL-based resource allocation
algorithm for the multi-UAV networks could achieve a tradeoff between the information exchange
overhead and system performance.

%

\subsection{Distributed and MA-Based Methods}

We proceed by surveying ML distributed methods that can be applied for resource allocation in UAV wireless networks.

Zeng et al. \cite{Zeng} used a distributed swarm approach for the power control and scheduling of UAV flocks. In particular, they used distributed federated learning (FL) algorithms within a UAV swarm that consisted of a leading UAV and several following UAVs. In their study, the UAVs trained their ability to share their local FL with the flock leader.

Chen et al. \cite{chenGT} considered the problem of spectrum
access in multi-UAV networks using a game theoretic perspective. They formulated the demand-aware joint channel slot selection problem as a weighted interference mitigation game, and subsequently designed a utility function considering the features of the multi-UAV network; for example, certain rewards based on the channel and slot selection. They proved that the formulated game was an exact potential game with at least one pure-strategy Nash equilibrium.

Thereafter,
they applied the distributed log-linear algorithm to
achieve the Nash equilibrium, and proposed a low-complexity and realistic
channel and slot initialization scheme for the UAVs to accelerate the convergence. They demonstrated by simulation that 
the network utility and aggregate interference level that were achieved
from the learning solution were very close to the best Nash equilibrium.


Liu et al. \cite{LiuPriceBased} studied the problem of downlink power allocation in UAV-based wireless networks with ABSs. They proposed a price-based optimal power allocation scheme and modeled the interaction between the UAVs and ground users as a Stackelberg game. As the leaders of the game, the UAVs selected the optimal power price to maximize their own revenue. Each ground user that belonged to the networks selected the optimal power strategy to maximize its own utility. To solve the equilibrium program with the equilibrium constraint problem of the Stackelberg game, they investigated the lower equilibrium of the ground users and proposed a distributed iterative algorithm to obtain the equilibrium of the lower game.

A distributed power control problem in a dense UAV network was formulated as a mean-field game (MFG) by Zhang et al. \cite{ZhangLi2}. They proposed a two-stage
scheme to alleviate the congestion of date traffic and interference effects. In the proposed scheme, a high-altitude platform (HAP) performs semi-persistent scheduling and allocates time frequency resources in a non-orthogonal manner, while the UAVs autonomously perform distributed power control. They formulated the
centralized resource allocation problem as a roommate matching problem and developed a novel time slot allocation algorithm to
solve it. Thereafter, the distributed power control of massive UAVs
was formulated as an MFG, which was solved
based on the finite difference method. The simulation results demonstrated
that the proposed scheme could significantly improve the reliability of
the communication in dense UAV networks.

In a recent study, Zhang et al. \cite{ZhangPowerControl}
proposed a potential game-based power control algorithm to optimize the power allocation of the UAVs. The power control problem was proven to be an exact potential game in which the Nash equilibrium was reached. First, the rule of determining neighbors was defined by fixing a threshold. Thereafter, the power strategies were updated continuously until convergence was
reached and the final optimal power allocation result was obtained.
They proposed a clustering algorithm by means of affinity propagation clustering (APC). The UAVs and their associated users were clustered into groups according to the mutual interference, whereby UAVs who potentially received severe interference from one another had a stronger tendency to join the same cluster. The APC also provided an unsupervised learning technique to determine the number of clusters and CHs to reduce interference automatically.

Table~\ref{table:ML:RA} summarizes the different studies, with their main goal, ML method that was used to achieve the main goal, how the method was tested and the validation thereof, and whether the project code can be downloaded.

\begin{table}
\caption{Summary of ML-based resource allocation methods (C = centralized, D = distributed) }
\label{table:ML:RA}
\begin{center}
\begin{tabular}{ | l | l | l | l | l | l |}\hline
Publication  & Resources & Model Objective & Learning Method & Approach & UAV\\ \hline
Sun et al. \cite{SunLearning} & Power control & throughput  & Deep learning & C & $\times$ \\ \hline 
Cui et al. \cite{CuiSpatial} &Link scheduling & Deep learning & C & V \\ \hline  \hline 
Zeng et al. \cite{Zeng}&  Power allocation &  Convergence rate & Distributed    &  D & V \\ 
&   and scheduling &      & federated learning &   &\\  \hline
Matthiesen  &  Power control & energy efficiency &Deep learning & C  & $\times$\\ 
et al. \cite{Matthiesen2} &   &    &  &    &\\ \hline
Ahmed et al. \cite{AhmedDeepRL} &  Resource allocation & throughput &Deep learning & C &$\times$\\ \hline
Chen et al. \cite{Chen2}&    User association,  & max stable   &  Liquid state  & D & V    \\ 
 &  spectrum allocation   & Queues &    Machine  & & \\ 
 &   &  and content    &    &       & \\   
 &   &  caching   &    &       & \\ \hline  
Lee at el. \cite{LeeKim} & Power control & Spectral efficiency/ & CNN & C/D&$\times$\\ 
  &  & energy efficiency &   & &\\ \hline
Zhang et al. \cite{Zhang}&   Power control & throughput  &RL &  C   & V  \\ \hline
Hu et al. \cite{Hu}&    User association,  &  maximize  &  RL& D   & V \\ 
 &      power management,   &  successful  &     & &\\ 
 &    subchannel alloc.   &   transmissions &  &     & \\ \hline \hline
Cao et al. \cite{Cao}& Channel and power  & Energy efficiency &  DRL  & C & V       \\ \hline
Ciu et al. \cite{Cui,CuiLiuML} & Dynamic resources &Long-term rewards & MA RL & D & V \\ \hline
Zeng et al. \cite{Zeng} & Power control, &  coverage &Distributed &  D & V \\ 
 &  scheduling & &federal &  &  \\ \hline
 &  scheduling & & learning &  &  \\ \hline
Chen et al. \cite{chenGT}&    Spectrum access & stability  &Log-linear    & D   & V  \\
 &      &  & algorithm to  &    & \\ 
 &      &  &  achieve NE  &    & \\ \hline
Liu et al.   &  Power allocation & Maximize  &pricing &  D & V \\ 
 \cite{LiuPriceBased} &      &  own values  & &    &  \\ \hline
Zhang et al.    & Power control & reliability  & Finite difference  & C+D & V \\  
  \cite{ZhangLi2}  &  &    & method  &   & \\ \hline
  Zhang et al. \cite{ZhangPowerControl} & Power control & Data traffic & APC clustering & C & V \\ \hline

\hline
\end{tabular}
\end{center}
\end{table}

\section{ML Methods for Task Allocation in UAV Flocks}
\label{section:allocation}
The problem of multi-type UAV task allocation involves determining the
best combination of UAVs to cover certain targets.


Huang et al. \cite{Huang2}  considered the multi-type UAV task allocation problem, which dealt with the question of how to determine the
best combination of UAVs to cover certain targets. They assumed that certain amounts of resources were required by the
targets and that the UAV formations had to meet the resource
demands of the targets to be qualified candidates, which
we refer to as resource constraints.
They presented a novel multi-type UAV coordinated task allocation method based on cross-entropy.
This study focused on the task allocation situation in which different types of UAVs were assigned to several tasks and the tasks required certain resources. The cross-entropy method obtained random samples from the candidate solutions and then used them to update the allocation probability matrix.

In \cite{Pohl}, the authors presented an algorithm for the routing of multiple UAVs and UAV swarms
to a set of locations while meeting the constraints of the time on target, total mission time, enemy radar avoidance, and total
path cost optimization. They aimed to develop a massive array of autonomous UAVs that were capable
of working together towards a common goal. They used a combinatorics
problem known as the vehicle routing problem with time windows (VRPTW) to model the routing situation of the UAVs.

In \cite{MuozMorera2016}, a planning approach for a platform composed of multiple heterogeneous unmanned aerial systems (UASs) was
presented. The research activities were
focused on the interoperability, task allocation, and task
planning problems within the system.
A consolidated planning technique was applied to generate low-level plans from a mission described in C-BML automatically,
thereby filling the gap between the interoperability and automatic
plan generation for missions in which multiple heterogeneous aerial platforms, both simulated and real, were involved. The architecture integrated symbolic, geometric,
and task allocation planners.
The approach was tested in missions
involving multiple surveillance and 3D map generation
tasks with a team of simulated and real UASs.

Hu et al. \cite{Hu2}
presented a new architecture for UAV clustering to enable efficient
multi-modal, multi-task offloading. They considered the problem in which the user distribution changes dynamically
in real time. Thus, the deployment of static/fixed edge computing nodes in
state-of-the-art networks failed to meet their computing
demands.
They constructed a model known as UAV-M3T for vehicular VR/AR gaming. The UAV-M3T architecture used AI-based decision making for the collaborative optimization of the UAV team and network resources to improve the task performance and resource efficiency
of the UAVs.
They proposed an AI-based decision-making framework to facilitate the UAV cooperation and joint optimization
of the computing, caching, and communication resources. In
this framework, the deployment of UAV clusters both in
advance, based on historical data mining, and in real time, based on real-time perception, were considered.


\subsection{Reinforcement Learning Methods}
\label{subsection:task:RL}
In \cite{Shyalika}, a review that focused on RL techniques for dynamic task scheduling was presented. The results of the study were addressed by means of the state of the art in RL used in dynamic task scheduling and a comparative review of those techniques. Although it did not focus on UAVs, it was connected to several studies in that area.

Shamsoshoara et al. \cite{Shamsoshoara} studied the problem of the limited spectrum in UAV networks and suggested a relay-based cooperative spectrum leasing scenario 
in which a group of UAVs in the network cooperatively forwarded data packets to a ground primary user for spectrum access.
Each UAV either joined a relaying group to provide a relaying service for the or performed data transmission to the UAV fusion center.
MARL was used for the task allocation among UAVs and simulation results were presented to
verify the convergence to the optimal solution.
The proposed method is a general form of the Q-learning algorithm for a distributed MA environment.

Kim and Morrison \cite{Kim}
jointly determined the system resources (system design), task allocation, and waypoint selection in a stochastic context. They
formulated the problem as an MDP and employed DRL to obtain state-based decisions. Numerical studies were conducted to assess the performance of the proposed approach.

The study of \cite{Zhao2019588} also used the MDP. These authors presented a Q-learning-based fast task allocation (FTA) algorithm through neural network approximation and prioritized experience replay, which
effectively offloaded the online computation to an offline learning procedure. Specifically, in the proposed
approach, a Q network was developed that encoded the allocation rules. The Q network 
could handle totally different tasks by
learning the allocation rules. The task allocation problem for heterogeneous UAVs in the presence of environmental uncertainty was formulated as an MDP. Subsequently, a Q-learning-based FTA algorithm using neural network approximation and prioritized experience replay was presented.

The study of \cite{YANG2019140} proposed a task-scheduling algorithm based on RL, which enabled the UAV to adjust its task strategy automatically and dynamically
using its calculation of the task performance efficiency.
A decentralized networking protocol
was presented to coordinate the movement of UAVs and to achieve real-time networking of UAV clusters.
The expansion strategy was applied to solve the problem of UAV networking in the initial state
and DRL was employed to solve the dynamic allocation problem of the wireless channel, so as to optimize the time delay of the UAV data transmission.
Finally, an example was provided to introduce the application of the above methods to solve the problem of UAV cluster task
scheduling.

In \cite{Bouhamed}, a spatiotemporal scheduling framework for autonomous UAVs using RL was presented. The framework enabled UAVs to determine their schedules autonomously to cover the maximum pre-scheduled events that were spatially and temporally distributed in a given geographical area and over a pre-determined time horizon. The designed framework had the ability to update the planned schedules in the case of unexpected emergency events. The UAVs were trained using the Q-learning algorithm to determine an effective scheduling plan.

\subsection{Bio Inspired Algorithms}
In \cite{KURDI2018110}, a bio-inspired approach for efficient task allocation was presented. The objective was to provide each UAV with the capability to make independent task allocation decisions to ensure autonomous control, an efficient running time, and the mean time to search and rescue a survivor.
They developed a new locust-inspired heuristic with explicit mapping of the algorithm components to the existing biological locust colony, and the proposed algorithm was customized to the task allocation
problem.
The proposed approach was inspired by the nature of the locust species, and their autonomous and elastic behavior in response to inside and outside stimuli.
The experimental results demonstrated the superiority of
the performance of LIAM compared to the well-established MUTA
heuristics, including the auction-based, max-sum, ACO, and opportunistic task allocation (OTA) algorithms,
where it successfully maintained a significantly higher throughput
as well as lower task completion and execution times.

In \cite{WANG2018339},
 a novel multi-UAV reconnaissance task allocation model was presented that considered the heterogeneity of targets as well as the constraints of the UAVs and targets. The optimization objective was to minimize the weighted sum of the task execution time and total UAV consumption. A modified GA was developed to solve the extended multiple Dubins traveling salesman problem (MDTSP). Double-chromosome encoding was used to describe the allocation results of targets to UAVs. Opposition-based learning and multiple mutation operators were employed to improve the optimality and convergence efficiency. The effectiveness of the algorithm was validated by numerical experiments on scenarios of different scales.

In \cite{Hu7347860}, a hierarchical method was presented to solve the task assignment problem for multiple UAV teams.
Tasks involving multiple UAVs formed into
several teams that cooperatively attacked numerous ground
targets were considered. In this scenario, the targets had to be
pre-assigned to the UAVs.
A hierarchical solution framework was developed to divide the task assignment problem into three subproblems: target clustering, cluster allocation, and target
assignment.

First, the targets were clustered based on
their geographic positions. Thereafter, the clusters were assigned
to different UAV teams, with one cluster per team, using the Hungarian algorithm. Finally, the targets were assigned to different team members for each team using an improved ACO (IACO) algorithm, the ant windows of which were adaptive to accelerate the convergence.

Through this process, all UAVs were assigned
targets along with the visit sequence.
The problem was solved by means of clustering algorithms and integer linear programming, using
a mixed integer linear programming model and the IACO. 
The ACO
 algorithm is a combinatorial optimization algorithm that
simulates the foraging process of ants, which is performed using chemical signals known as pheromones.

In \cite{Kurdi}, the author introduced a bio-inspired algorithm for task allocation in multi-UAV search and rescue missions based on locust behavior. The locust species was selected as it represents an extreme example of adaptive control and high plasticity to respond to internal and external stimuli. To evaluate the proposed algorithm, a strictly controlled experimental framework was designed to demonstrate the potential role of the proposed algorithm in such missions by providing a high net throughput and an efficient communication scheme.

In \cite{Ye20}, the cooperative multiple task assignment problem (CMTAP) of heterogeneous fixed-wing UAVs performing the Suppression of Enemy Air Defense (SEAD) mission against multiple ground stationary targets was discussed. The study presented a modified GA with a multi-type gene chromosome encoding strategy.
First, the multi-type gene encoding scheme was introduced to generate feasible chromosomes that satisfied the UAV capability, task coupling, and task precedence constraints. Thereafter, the Dubins car model was adopted to calculate the mission execution time (the objective function of the CMTAP model) of each chromosome and to make each chromosome conform to the UAV maneuverability constraint.

In \cite{Luo17}, the process of allocating pesticide spraying tasks by multiple UAVs was investigated. A model was proposed to produce a satisfactory allocation scenario for pesticide spraying and to provide the flight route for each UAV. The model extended the distance between two points to the Dubins path distance and dynamically determined the time windows of the pesticide spraying according to the temperature conditions at the time that the UAVs performed the tasks.
The study proposed a GA to solve the aforementioned problem and provided the encoding, crossover, and mutation methods in the algorithm.

The study of \cite{KURDI2019105643} proposed a bacteria-inspired heuristic for the efficient distribution of tasks among deployed UAVs. The use of multi-UAVs is a promising concept for combating the spread of the red palm weevil (RPW) in palm plantations. The proposed bacteria-inspired heuristic was used to resolve the multi-UAV task allocation problem when combating RPW infestation. The performance of the proposed algorithm was benchmarked in simulated detect-and-treat missions against three long-standing multi-UAV task allocation strategies, namely opportunistic task allocation, the auction-based scheme, and the max-sum algorithm, as well as a recently introduced locust-inspired algorithm for the allocation of multi-UAVs.



\begin{table}
\label{table:ML:allocation}
\caption{Task Allocation}
\begin{center}
  \begin{tabular}{ | l | l | l | l | }\hline
 Publication & Challenge & Optimization & ML   \\ 
   &   & Criteria &  Method \\ \hline 
Huang et al. \cite{Huang2} &Coordinated & team rewards & Cross-entropy\\
 &  task allocation- &   &based algorithm \\ \hline
Pohl and Lamont \cite{Pohl} & UAV mission planning& path length & PSO, GA\\ \hline
Munoz-Morera   & Task allocation & cost & planning \\ 
  et al. \cite{MuozMorera2016} & & & \\ \hline
Hu et al. \cite{Hu2}& Multi-modal, multi-task  & min delay & RNN \\ 
 &   task offloading & &   \\ \hline \hline
Shyalika et al. \cite{Shyalika}& RL techniques  & $\times$ & RL\\ 
& for task scheduling & &\\ \hline
Shamsoshoara et al. & UAV partition & communication & Distributed \\
\cite{Shamsoshoara} & and task allocation & rate & Q-learning \\ \hline
Kim and Morrison \cite{Kim} &  Waypoint selection & reward & DRL \\ \hline
Zhao et al. \cite{Zhao2019588} & Task allocation for heterogeneous &Reward and  & Q-learning \\
 &UAVs in uncertain environment & coverage&  \\\hline

Yang et al. \cite{YANG2019140} &Minimizing sum power for   & Transmission & DRL \\
  &UAV-enabled MEC network  & delay &   \\\hline
Bouhamed et al. \cite{Bouhamed} & Spatiotemporal scheduling   & coverage & Q-learning \\ 
 \hline \hline
Kurdi et al. \cite{KURDI2018110} & UAV allocation to survivors  & Throughput  & Locust inspired   \\ \hline
 Wang et al. \cite{WANG2018339} & Task allocation for & path length & GA  \\
 & heterogeneous targets& & \\ \hline
 Hu et al. \cite{Hu7347860} & Task allocation & Overall reward & Clustering +    \\
&  for multiple teams &   &   Hungarian alg. + \\
&  &   &    ACO \\\hline
Kurdi et al. \cite{Kurdi} & Coordinated task allocation & Throughput  & Locust inspired  \\ \hline
Kim et al.  & Resource and task allocation  & Reward and cost & DRL\\ \hline
Fang et al. \cite{Ye20} & Cooperative task assignment   & Mission execution & GA \\ 
  &   against ground stationary targets & time  &  \\ \hline
Luo et al. \cite{Luo17} & Pesticide spraying task allocation & Total profit  & GA \\ \hline
 Kurdi et al. \cite{KURDI2019105643} & Task allocation  & & Bacteria-inspired  \\ 
  &  among deployed UAVs &Net  throughput & heuristic \\ \hline

  \end{tabular}
   
\end{center}
\end{table}


\section{Open Issues}
\label{section:conclusion}

Several open issues can be considered in the different challenges involving UAV flocks.
First, most studies have suggested one or two ML-based methods to solve the flock challenge, and subsequently compared the performance of their methods to classical algorithmic solutions presented in the literature. However, a comparison of different ML methods has rarely been conducted, and this may be important to enable the correct mapping of an ML method to each specific challenge.

Another important issue is how to reflect the effects of a decision on flock management in the future system state. These effects can be incorporated by using RL and DRL methods, probably combined with other ML methods.

The combination of more than one DRL method may also be efficient for different real problems.
For example, a CNN and LSTM can be combined to achieve superior solutions when considering sequences of several prediction problems over a time series given coverage maps. Moreover, DRL and CNN or LSTM can be combined to achieve better decision making in complex spatial or sequential models.

Distributed algorithms have been widely suggested for flock management. However, when considering a UAV that performs its tasks alone without interacting in a flow (hereafter referred to as a standalone UAV), the use of distributed algorithms is rare. A standalone UAV behaves in a manner that is best for itself, and it will cooperate with other UAVs only if this will improve its own situation. Thus, any mechanism that is proposed for such constellations should be proven to be stable. DRL methods can be used in such situations, considering self-learning by each UAV or by each UAV group, but it is not clear whether centralized learning can be implemented by any standalone CH or cluster presenter, given the fact that it may learn a model that is biased for its own profit rather than for the benefit of the entire flock or system.

The geographical aspects of UAV motion and communication, such as the localization, deployment, task allocation, and resource allocation, can include the altitude dimension, which may affect the distance between UAVs, and the solution efficiency can be improved when considering the ability to improve the use of the altitude parameter.

Another challenge that will affect future developments in this area is the need to obtain real-world-based standard flock databases, with data regarding their motions and actions, for use as comparative databases for ML training when solving algorithm challenges relating to flock formation and maintenance. Databases for standard UAV motions and actions may also be useful to compare different ML methods, which is already the case in other domains in which ML is widely used, such as NLP and machine vision.

Finally, the abilities of different deep learning schemes make it possible to take a harmonic approach, and to consider all aspects of flock management and use. This enables the provision of  an overall ML solution that solves all of the involved challenges, including flock formation, flock maintenance, resource allocation, task allocation, and UAV deployment, simultaneously. Such a harmonic view may be efficient and can provide a real-time, rapid solution to several applications of UAV flocks, which are expected to become increasingly popular in the near future, as the cost of UAV-based technology becomes lower and their use becomes more crucial.


\section{Conclusions}

This paper has surveyed several complex issues relating to UAV flock formation, maintenance, and related challenges, which can be efficiently handled using ML methods. We have provided a review that is as current as possible, and in each section we also discussed relevant open issues.  
We hope that our study will be helpful for researchers and developers who wish to apply ML methods for solving challenges relating to UAV flocks, and in particular, regarding FANETs.

The technological advancements in the production of UAVs, together with the need to manage them by grouping them into clusters, offers enormous potential in various fields, including commercial, industrial, agricultural, and security applications. Moreover, several challenges arise in determining the correct manner in which to manage UAV clusters.
ML methods can enable rapid and efficient solutions to these challenges in real time for problems that can be considered in advance, so that the learning system can be trained in a manner that will make it possible to handle the various challenges in real time effectively. Such challenges appear in several main areas: flock formation and management; flock deployment, coordination, and navigation; inter-flock and intra-flock communication resource allocation; and finally, task allocation in flocks that are responsible for handling predefined tasks.
To the best of our knowledge, this survey is the first attempt to address the range of issues associated with the creation, management, and maintenance of UAV clusters comprehensively, with particular emphasis on machine-based solutions, to provide an effective and rapid response to these important issues.

\bibliographystyle{plain}
\bibliography{ref}

\begin{thebibliography}{100}

\bibitem{Koushik}
F.~Hu A.~M.~Koushik and S.~Kumar.
\newblock Deep q-learning-based node positioning for throughput-optimal
  communications in dynamic uav swarm network.
\newblock {\em IEEE Transactions on Cognitive Communications and Networking},
  5(3):554--566, 2019.

\bibitem{Carrio}
Alejandro Rodríguez~Ramos Adrian~Carrio, Carlos Sampedro~Pérez and Pascual
  Campoy.
\newblock A review of deep learning methods and applications for unmanned
  aerial vehicles.
\newblock {\em Journal of Sensors}, pages 1--13, 2017.

\bibitem{AftabZhang}
F.~{Aftab}, A.~{Khan}, and Z.~{Zhang}.
\newblock Hybrid self-organized clustering scheme for drone based cognitive
  internet of things.
\newblock {\em IEEE Access}, 7:56217--56227, 2019.

\bibitem{AgarwalNPH}
P.~Agarwal and C.~Procopiuc.
\newblock Exact and approximation algorithms for clustering.
\newblock In {\em Proceedings of the Ninth Annual A CM-SIAM Symposium on
  Discrete Algorithms}, pages 658--667, 1999.

\bibitem{AhmedDeepRL}
K.~I. {Ahmed}, H.~{Tabassum}, and E.~{Hossain}.
\newblock Deep learning for radio resource allocation in multi-cell networks.
\newblock {\em IEEE Network}, 33(6):188--195, 2019.

\bibitem{AlTurjman}
Fadi Al-Turjman and Hadi Zahmatkesh.
\newblock {\em A Comprehensive Review on the Use of AI in UAV Communications:
  Enabling Technologies, Applications, and Challenges}, pages 1--26.
\newblock Springer, 2020.

\bibitem{Alsamhi32}
S.H. Alsamhi, O.~Ma, and M.S. Ansari.
\newblock Survey on artificial intelligence based techniques for emerging
  robotic communication.
\newblock {\em Telecommun. Syst.}, 72:1--21, 2019.

\bibitem{anicho}
Ogbonnaya Anicho, Philip~B Charlesworth, Gurvinder~S Baicher, Atulya Nagar, and
  Neil Buckley.
\newblock Comparative study for coordinating multiple unmanned haps for
  communications area coverage.
\newblock In {\em 2019 International Conference on Unmanned Aircraft Systems
  (ICUAS)}, 2019.

\bibitem{ArafatClustering}
M.~Y. Arafat and S.~Moh.
\newblock Localization and clustering based on swarm intelligence in uav
  networks for emergency communications.
\newblock {\em IEEE Internet of Things Journal}, 6(5), 2019.

\bibitem{Sargolzaei}
Carl D.~Crane1 Arman~Sargolzaei, Alireza~Abbaspour.
\newblock {\em Control of Cooperative Unmanned Aerial Vehicles: Review of
  Applications, Challenges, and Algorithms}, volume 1123, pages 229--255.
\newblock Springer, 2020.

\bibitem{KirilPowerControl}
Rina Azoulay, Kiril Danilchenko, Yoram Haddad, and Shulamit Reches.
\newblock Transmit power control by deep neural network in tdma-based ad-hoc
  clusters.
\newblock In {\em IWCMC}, 2021.

\bibitem{AzoulayReches}
Rina Azoulay and Shulamit Reches.
\newblock Uav flocks forming for crowded flight environments.
\newblock In {\em ICAART}, 2019.

\bibitem{AzoulayRechesJournal}
Rina Azoulay and Shulamit Reches.
\newblock Flocks formation model for self-interested uavs.
\newblock {\em Intelligent Service Robotics}, 2021.

\bibitem{Bayerlein}
H.~{Bayerlein}, P.~{De Kerret}, and D.~{Gesbert}.
\newblock Trajectory optimization for autonomous flying base station via
  reinforcement learning.
\newblock In {\em 2018 IEEE 19th International Workshop on Signal Processing
  Advances in Wireless Communications (SPAWC)}, pages 1--5, 2018.

\bibitem{beaver2020overview}
Logan~E. Beaver and Andreas~A. Malikopoulos.
\newblock An overview on optimal flocking, 2020.

\bibitem{Bithas}
Petros~S. Bithas, Emmanouel~T. Michailidis, Nikolaos Nomikos, Demosthenes
  Vouyioukas, and Athanasios~G. Kanatas.
\newblock A survey on machine-learning techniques for uav-based communications.
\newblock {\em Sensors}, 19(23), 2019.

\bibitem{Bouhamed}
O.~{Bouhamed}, H.~{Ghazzai}, H.~{Besbes}, and Y.~{Massoud}.
\newblock A generic spatiotemporal scheduling for autonomous uavs: A
  reinforcement learning-based approach.
\newblock {\em IEEE Open Journal of Vehicular Technology}, 1:93--106, 2020.

\bibitem{LiuNavigation}
X.~Gao C.~H.~Liu, X.~Ma and J.~Tang.
\newblock Distributed energy-efficient multi-uav navigation for long-term
  communication coverage by deep reinforcement learning.
\newblock {\em IEEE Transactions on Mobile Computing}, 19(6):1274--1285, 2020.

\bibitem{Cekmez}
U.~{Cekmez}, M.~{Ozsiginan}, and O.~K. {Sahingoz}.
\newblock Multi colony ant optimization for uav path planning with obstacle
  avoidance.
\newblock In {\em 2016 International Conference on Unmanned Aircraft Systems
  (ICUAS)}, pages 47--52, 2016.

\bibitem{ChallitaInterference1}
U.~{Challita}, W.~{Saad}, and C.~{Bettstetter}.
\newblock Deep reinforcement learning for interference-aware path planning of
  cellular-connected uavs.
\newblock In {\em 2018 IEEE International Conference on Communications (ICC)},
  pages 1--7, 2018.

\bibitem{ChallitaDRL}
Ursula Challita, Walid Saad, and Christian Bettstetter.
\newblock Cellular-connected uavs over 5g: Deep reinforcement learning for
  interference management, 2018.

\bibitem{Zhang2}
Paul~Patras Chaoyun~Zhang and Hamed Haddadi.
\newblock Deep learning in mobile and wireless networking: A survey.
\newblock {\em IEEE Communications surveys and Tutorials}, 2019.

\bibitem{chenGT}
J.~Chen, Q.~Wu, Y.~Xu, Y.~Zhang, and Y.~Yang.
\newblock Distributed demand-aware channel-slot selection for multi-uav
  networks: A game-theoretic learning approach.
\newblock {\em IEEE Access}, 6, 2018.

\bibitem{Chen}
M.~{Chen}, U.~{Challita}, W.~{Saad}, C.~{Yin}, and M.~{Debbah}.
\newblock Artificial neural networks-based machine learning for wireless
  networks: A tutorial.
\newblock {\em IEEE Communications Surveys Tutorials}, 21(4):3039--3071, 2019.

\bibitem{ChenCachingSky}
M.~Chen, M.~Mozaffari, W.~Saad, C.~Yin, M.~Debbah, and C.~S. Hong.
\newblock Caching in the sky: Proactive deployment of cache-enabled unmanned
  aerial vehicles for optimized quality-of-experience.
\newblock {\em IEEE Journal on Selected Areas in Communications},
  35(5):1046--1061, 2017.

\bibitem{Chen2}
M.~{Chen}, W.~{Saad}, and C.~{Yin}.
\newblock Liquid state machine learning for resource and cache management in
  lte-u unmanned aerial vehicle (uav) networks.
\newblock {\em IEEE Transactions on Wireless Communications}, 18(3):1504--1517,
  2019.

\bibitem{XiJun}
Xi~Chen, Jun Tang, and Songyang Lao.
\newblock Review of unmanned aerial vehicle swarm communication architectures
  and routing protocols.
\newblock {\em Appl. Sci.}, 10(10):3661, 2020.

\bibitem{ChungSurvey}
S.~{Chung}, A.~A. {Paranjape}, P.~{Dames}, S.~{Shen}, and V.~{Kumar}.
\newblock A survey on aerial swarm robotics.
\newblock {\em IEEE Transactions on Robotics}, 34(4):837--855, 2018.

\bibitem{Collotta}
M.~{Collotta}, G.~{Pau}, and V.~{Maniscalco}.
\newblock A fuzzy logic approach by using particle swarm optimization for
  effective energy management in iwsns.
\newblock {\em IEEE Transactions on Industrial Electronics}, 64(12):9496--9506,
  2017.

\bibitem{Coppola}
Mario Coppola, Kimberly~N. McGuire, Christophe~De Wagter, and Guido C. H.~E.
  de~Croon.
\newblock A survey on swarming with micro air vehicles: Fundamental challenges
  and constraints.
\newblock {\em Front. Robotics and AI}, 2020.

\bibitem{CuiLiuML}
J.~{Cui}, Y.~{Liu}, and A.~{Nallanathan}.
\newblock The application of multi-agent reinforcement learning in uav
  networks.
\newblock In {\em 2019 IEEE International Conference on Communications
  Workshops (ICC Workshops)}, pages 1--6, 2019.

\bibitem{Cui}
J.~{Cui}, Y.~{Liu}, and A.~{Nallanathan}.
\newblock Multi-agent reinforcement learning-based resource allocation for uav
  networks.
\newblock {\em IEEE Transactions on Wireless Communications}, 19(2):729--743,
  2020.

\bibitem{CuiSpatial}
W.~{Cui}, K.~{Shen}, and W.~{Yu}.
\newblock Spatial deep learning for wireless scheduling.
\newblock {\em IEEE Journal on Selected Areas in Communications},
  37(6):1248--1261, 2019.

\bibitem{CynaraDistributed}
{Cynara Wu} and D.~P. {Bertsekas}.
\newblock Distributed power control algorithms for wireless networks.
\newblock {\em IEEE Transactions on Vehicular Technology}, 50(2):504--514,
  2001.

\bibitem{DaiDeploy}
H.~{Dai}, H.~{Zhang}, M.~{Hua}, C.~{Li}, Y.~{Huang}, and B.~{Wang}.
\newblock How to deploy multiple uavs for providing communication service in an
  unknown region.
\newblock {\em IEEE Wireless Communications Letters}, 8(4):1276--1279, 2019.

\bibitem{Dai}
Haibo Dai, Haiyang Zhang, Baoyun Wang, and Luxi Yang.
\newblock The multi-objective deployment optimization of uav-mounted
  cache-enabled base stations.
\newblock {\em Physical Communication}, 34:114--120, 2019.

\bibitem{Fotouhi}
A.~{Fotouhi}, H.~{Qiang}, M.~{Ding}, M.~{Hassan}, L.~G. {Giordano},
  A.~{Garcia-Rodriguez}, and J.~{Yuan}.
\newblock Survey on uav cellular communications: Practical aspects,
  standardization advancements, regulation, and security challenges.
\newblock {\em IEEE Communications Surveys Tutorials}, 21(4):3417--3442, 2019.

\bibitem{Ganesan}
R.~Ganesan, X.M. Raajini, A.~Nayyar, P.~Sanjeevikumar, E.~Hossain, and A.H.
  Ertas.
\newblock Bold: Bio-inspired optimized leader election for multiple drones.
\newblock {\em Sensors}, 20, 2020.

\bibitem{Govindaraj}
S.~Govindaraj and S.N Deepa.
\newblock Network energy optimization of iots in wireless sensor networks using
  capsule neural network learning model.
\newblock {\em Wireless Pers Commun}, 115:2415--2436, 2020.

\bibitem{Hayat16}
S.~{Hayat}, E.~{Yanmaz}, and R.~{Muzaffar}.
\newblock Survey on unmanned aerial vehicle networks for civil applications: A
  communications viewpoint.
\newblock {\em IEEE Communications Surveys Tutorials}, 18(4):2624--2661, 2016.

\bibitem{HeDS}
H.~{He}, Z.~{Zhu}, and E.~{Makinen}.
\newblock A neural network model to minimize the connected dominating set for
  self-configuration of wireless sensor networks.
\newblock {\em IEEE Transactions on Neural Networks}, 20(6):973--982, 2009.

\bibitem{HentatiKrichen}
A.~I. {Hentati}, L.~{Krichen}, M.~{Fourati}, and L.~C. {Fourati}.
\newblock Simulation tools, environments and frameworks for uav systems
  performance analysis.
\newblock In {\em 2018 14th International Wireless Communications Mobile
  Computing Conference (IWCMC)}, pages 1495--1500, 2018.

\bibitem{Idriss}
Aicha~Idriss Hentati and Lamia~Chaari Fourati.
\newblock Comprehensive survey of uavs communication networks.
\newblock {\em Computer Standards and Interfaces}, 72, 2020.

\bibitem{Hu}
J.~{Hu}, H.~{Zhang}, L.~{Song}, Z.~{Han}, and H.~V. {Poor}.
\newblock Reinforcement learning for a cellular internet of uavs: Protocol
  design, trajectory control, and resource management.
\newblock {\em IEEE Wireless Communications}, 27(1):116--123, 2020.

\bibitem{Hu2}
L.~Hu, Y.~Tian, J.~Yang, T.~Taleb, L.~Xiang, and Y.~Hao.
\newblock Ready player one: Uav-clustering-based multi-task offloading for
  vehicular vr/ar gaming.
\newblock {\em IEEE Network}, 33(3):42--48, 2019.

\bibitem{Hu7347860}
X.~{Hu}, H.~{Ma}, Q.~{Ye}, and H.~{Luo}.
\newblock Hierarchical method of task assignment for multiple cooperating uav
  teams.
\newblock {\em Journal of Systems Engineering and Electronics},
  26(5):1000--1009, 2015.

\bibitem{Hung}
S.~Hung and S.~N. Givigi.
\newblock A q-learning approach to flocking with uavs in a stochastic
  environment.
\newblock {\em IEEE Transactions on Cybernetics}, 47(1):186--197, 2017.

\bibitem{Bekmezci}
Ozgur Koray~Sahingoz Ilker~Bekmezci and Samil Temel.
\newblock Flying ad-hoc networks (fanets): A survey.
\newblock {\em Ad Hoc Networks}, 11(3):1254--1270, 2013.

\bibitem{HuRLTrajectory}
H.~Zhang J.~Hu and L.~Song.
\newblock Reinforcement learning for decentralized trajectory design in
  cellular uav networks with sense-and-send protocol.
\newblock {\em IEEE Internet of Things Journal}, 6(4):6177--6189, 2019.

\bibitem{Jafari}
M.~Jafari, H~H.~Xu, and L.R.G. Carrillo.
\newblock A biologically inspired reinforcement learning based intelligent
  distributed flocking control for multi-agent systems in presence of uncertain
  system and dynamic environment.
\newblock {\em IFAC Journal of Systems and Control}, 13, 2020.

\bibitem{ZhangPowerControl}
Gang~Chuai Jinxi~Zhang and Weidong Gao.
\newblock Power control and clustering-based interference management for
  uav-assisted networks.
\newblock {\em Sensors}, 20(14), 2020.

\bibitem{KhalilNPH}
Elias~Boutros Khalil, Hanjun Dai, Yuyu Zhang, Bistra~N. Dilkina, and Le~Song.
\newblock Learning combinatorial optimization algorithms over graphs.
\newblock {\em Neural Information Processing Systems}, 2017.

\bibitem{Khamidehi}
B.~{Khamidehi} and E.~S. {Sousa}.
\newblock Reinforcement learning-based trajectory design for the aerial base
  stations.
\newblock In {\em 2019 IEEE 30th Annual International Symposium on Personal,
  Indoor and Mobile Radio Communications (PIMRC)}, pages 1--6, 2019.

\bibitem{Kim}
I.~Kim and J.~R. Morrison.
\newblock Learning based framework for joint task allocation and system design
  in stochastic multi-uav systems.
\newblock In {\em 2018 International Conference on Unmanned Aircraft Systems
  (ICUAS)}, pages 324--334, 2018.

\bibitem{Klaine}
P.V. Klaine, J.P.B. Nadas, and R.D. et~al. Souza.
\newblock Distributed drone base station positioning for emergency cellular
  networks using reinforcement learning.
\newblock {\em Cogn Comput}, 10(790–804), 2018.

\bibitem{Kouhdaragh}
Vahid Kouhdaragh, Francesco Verde, Giacinto Gelli, and Jamshid Abouei.
\newblock On the application of machine learning to the design of uav-based 5g
  radio access networks.
\newblock {\em Electronics}, 9, 2020.

\bibitem{KURDI2019105643}
Heba Kurdi, Munirah~F. AlDaood, Shiroq Al-Megren, Ebtesam Aloboud,
  Abdulrahman~S. Aldawood, and Kamal Youcef-Toumi.
\newblock Adaptive task allocation for multi-uav systems based on bacteria
  foraging behaviour.
\newblock {\em Applied Soft Computing}, 83:105643, 2019.

\bibitem{Kurdi}
Heba Kurdi, Jonathon How, and Guillermo Bautista.
\newblock Bio-inspired algorithm for task allocation in multi-uav search and
  rescue missions.
\newblock In {\em AIAA}, 01 2016.

\bibitem{KURDI2018110}
Heba~A. Kurdi, Ebtesam Aloboud, Maram Alalwan, Sarah Alhassan, Ebtehal
  Alotaibi, Guillermo Bautista, and Jonathan~P. How.
\newblock Autonomous task allocation for multi-uav systems based on the locust
  elastic behavior.
\newblock {\em Applied Soft Computing}, 71:110 -- 126, 2018.

\bibitem{Nguyen}
A.~Kortun L.~D.~Nguyen, K. K.~Nguyen and T.~Q. Duon.
\newblock Real-time deployment and resource allocation for distributed uav
  systems in disaster relief.
\newblock In {\em 2019 IEEE 20th International Workshop on Signal Processing
  Advances in Wireless Communications (SPAWC)}, pages 1--5, 2019.

\bibitem{Huang2}
H.~Qu L.~Huang and L.~Zuo.
\newblock Multi-type uavs cooperative task allocation under resource
  constraints.
\newblock {\em IEEE Access}, 6:17841--17850, 2018.

\bibitem{Latiff}
N.~M.~A. {Latiff}, C.~C. {Tsimenidis}, and B.~S. {Sharif}.
\newblock Energy-aware clustering for wireless sensor networks using particle
  swarm optimization.
\newblock In {\em 2007 IEEE 18th International Symposium on Personal, Indoor
  and Mobile Radio Communications}, pages 1--5, 2007.

\bibitem{LeeKim}
W.~{Lee}, M.~{Kim}, and D.~{Cho}.
\newblock Deep power control: Transmit power control scheme based on
  convolutional neural network.
\newblock {\em IEEE Communications Letters}, 22(6):1276--1279, 2018.

\bibitem{LiFeiZhang}
B.~{Li}, Z.~{Fei}, and Y.~{Zhang}.
\newblock Uav communications for 5g and beyond: Recent advances and future
  trends.
\newblock {\em IEEE Internet of Things Journal}, 6(2):2241--2263, 2019.

\bibitem{Liu}
C.~H. Liu, Z.~Chen, J.~Tang, J.~Xu, and C.~Piao.
\newblock Energy-efficient uav control for effective and fair communication
  coverage: A deep reinforcement learning approach.
\newblock {\em IEEE Journal on Selected Areas in Communications},
  36(9):2059–2070, 2018.

\bibitem{LiuWangFast}
J.~{Liu}, Q.~{Wang}, X.~{Li}, and W.~{Zhang}.
\newblock A fast deployment strategy for uav enabled network based on deep
  learning.
\newblock In {\em 2020 IEEE 31st Annual International Symposium on Personal,
  Indoor and Mobile Radio Communications}, pages 1--6, 2020.

\bibitem{LiuPriceBased}
X.~{Liu}, L.~{Li}, F.~{Yang}, X.~{Li}, W.~{Chen}, and W.~{Xu}.
\newblock Price-based power allocation for multi-uav enabled wireless networks.
\newblock In {\em 2019 28th Wireless and Optical Communications Conference
  (WOCC)}, pages 1--5, 2019.

\bibitem{LiuLiuClustering}
X.~{Liu}, Y.~{Liu}, and Y.~{Chen}.
\newblock Reinforcement learning in multiple-uav networks: Deployment and
  movement design.
\newblock {\em IEEE Transactions on Vehicular Technology}, 68(8):8036--8049,
  2019.

\bibitem{Long}
Nathan~K Long, Karl Sammut, Daniel Sgarioto, Matthew Garratt, and Hussein.
\newblock A comprehensive review of shepherding as a bio-inspired
  swarm-robotics guidance approach.
\newblock {\em IEEE Transactions On Emerging Topics In Computational
  Intelligence}, 21(4):3039--3071, 2020.

\bibitem{Luo17}
He~Luo, Yanqiu Niu, Moning Zhu, Xiaoxuan hu, and Huawei Ma.
\newblock Optimization of pesticide spraying tasks via multi-uavs using genetic
  algorithm.
\newblock {\em Mathematical Problems in Engineering}, 2017:1--16, 11 2017.

\bibitem{LuoZhang}
Z.~{Luo} and S.~{Zhang}.
\newblock Dynamic spectrum management: Complexity and duality.
\newblock {\em IEEE Journal of Selected Topics in Signal Processing},
  2(1):57--73, 2008.

\bibitem{Lyu2017}
J.~{Lyu}, Y.~{Zeng}, R.~{Zhang}, and T.~J. {Lim}.
\newblock Placement optimization of uav-mounted mobile base stations.
\newblock {\em IEEE Communications Letters}, 21(3):604--607, 2017.

\bibitem{Matthiesen2}
Bho Matthiesen, Alessio Zappone, Eduard Jorswieck, and Mérouane Debbah.
\newblock Deep learning for real-time energy-efficient power control in mobile
  networks.
\newblock In {\em IEEE 20th International Workshop on Signal Processing
  Advances in Wireless Communications (SPAWC)}, 2019.

\bibitem{Maza}
I.~Maza, F.~Caballero, J.~Capitan, J.~R.~Martínez de~Dios, and A.~Ollero.
\newblock Experimental results in multi-uav coordination for disaster
  management and civil security applications.
\newblock {\em Journal of Intelligent and Robotic Systems}, 61:1--4, 2011.

\bibitem{MazaAWARE}
I.~Maza, K.~Kondak, M.~Bernard, and A.~Ollero.
\newblock Multi-uav cooperation and control for load transportation and
  deployment.
\newblock {\em Journal of Intelligent and Robotic Systems}, 57:417–449, 2010.

\bibitem{Mirjalili}
S.~Mirjalili.
\newblock Dragonfly algorithm: A new meta-heuristic optimization technique for
  solving single-objective, discrete, and multi-objective problems.
\newblock {\em Neural Comput. Appl.}, 27:1053–1073, 2016.

\bibitem{Mozaffari}
M.~{Mozaffari}, W.~{Saad}, M.~{Bennis}, Y.~{Nam}, and M.~{Debbah}.
\newblock A tutorial on uavs for wireless networks: Applications, challenges,
  and open problems.
\newblock {\em IEEE Communications Surveys Tutorials}, 21(3):2334--2360, 2019.

\bibitem{MuozMorera2016}
Jorge Mu{\~n}oz-Morera, I.~Maza, F.~Caballero, and A.~Ollero.
\newblock Architecture for the automatic generation of plans for multiple uas
  from a generic mission description.
\newblock {\em Journal of Intelligent and Robotic Systems}, 84:493--509, 2016.

\bibitem{Olfati-Saber}
R.~{Olfati-Saber}.
\newblock Flocking for multi-agent dynamic systems: algorithms and theory.
\newblock {\em IEEE Transactions on Automatic Control}, 51(3):401--420, 2006.

\bibitem{Oubbati}
O.~S. {Oubbati}, M.~{Atiquzzaman}, P.~{Lorenz}, M.~H. {Tareque}, and M.~S.
  {Hossain}.
\newblock Routing in flying ad hoc networks: Survey, constraints, and future
  challenge perspectives.
\newblock {\em IEEE Access}, 7:81057--81105, 2019.

\bibitem{ParkLee}
Y.~M. {Park}, M.~{Lee}, and C.~S. {Hong}.
\newblock Multi-uavs collaboration system based on machine learning for
  throughput maximization.
\newblock In {\em 2019 20th Asia-Pacific Network Operations and Management
  Symposium (APNOMS)}, pages 1--4, 2019.

\bibitem{PengQ}
Jing Peng and Ronald~J. Williams.
\newblock Incremental multi-step q-learning.
\newblock {\em Machine Learning}, 22:283--290, 1996.

\bibitem{Boccadoro}
Domenico~Striccoli Pietro~Boccadoro and Luigi~Alfredo Grieco.
\newblock Internet of drones: a survey on communications, technologies,
  protocols, architectures and services.
\newblock {\em arXiv:2007.12611}, 2020.

\bibitem{Pohl}
A.~J. Pohl and G.~B. Lamont.
\newblock Multi-objective uav mission planning using evolutionary computation.
\newblock In {\em 2008 Winter Simulation Conference}, pages 1268--1279, 2008.

\bibitem{Kuila}
Prasanta K.~Jana Pratyay~Kuila.
\newblock Energy efficient clustering and routing algorithms for wireless
  sensor networks: Particle swarm optimization approach.
\newblock {\em Engineering Applications of Artificial Intelligence},
  33:127--140, 2014.

\bibitem{PanKMeans}
Pan Q., Ni~Q., Du~H., Yao Y., and Lv~Q.
\newblock An improved energy-aware cluster heads selection method for wireless
  sensor networks based on k-means and binary particle swarm optimization.
\newblock {\em Advances in Swarm Intelligence}, 8795, 2014.

\bibitem{Mao}
Qi~Hao Qian~Mao, Fei~Hu.
\newblock Deep learning for intelligent wireless networks: A comprehensive
  survey.
\newblock {\em IEEE Communications Surveys and Tutorials}, 20(4), 2018.

\bibitem{QieShi}
H.~{Qie}, D.~{Shi}, T.~{Shen}, X.~{Xu}, Y.~{Li}, and L.~{Wang}.
\newblock Joint optimization of multi-uav target assignment and path planning
  based on multi-agent reinforcement learning.
\newblock {\em IEEE Access}, 7:146264--146272, 2019.

\bibitem{Qie}
H.~{Qie}, D.~{Shi}, T.~{Shen}, X.~{Xu}, Y.~{Li}, and L.~{Wang}.
\newblock Joint optimization of multi-uav target assignment and path planning
  based on multi-agent reinforcement learning.
\newblock {\em IEEE Access}, 7:146264--146272, 2019.

\bibitem{QiuLyu2020}
J.~{Qiu}, J.~{Lyu}, and L.~{Fu}.
\newblock Placement optimization of aerial base stations with deep
  reinforcement learning.
\newblock In {\em ICC 2020 - 2020 IEEE International Conference on
  Communications (ICC)}, pages 1--6, 2020.

\bibitem{Quintero}
S.~A.~P. {Quintero}, G.~E. {Collins}, and J.~P. {Hespanha}.
\newblock Flocking with fixed-wing uavs for distributed sensing: A stochastic
  optimal control approach.
\newblock In {\em 2013 American Control Conference}, pages 2025--2031, 2013.

\bibitem{Dabbiru}
Vikranth~Dabbiru R. and Siddhabathula K.
\newblock Autonomous air traffic control - collision avoidance for uavs using
  mdp.
\newblock {\em International Journal of Computer Science and Information
  Technology and Security}, 6(1), 2016.

\bibitem{Reynolds}
Craig Reynolds.
\newblock Flocks, herds and schools: A distributed behavioral model.
\newblock {\em ACM SIGGRAPH Computer Graphics}, 1987.

\bibitem{Ragi2013}
Ragi S. and E.K.P. Chong.
\newblock Uav path planning in a dynamic environment via partially observable
  markov decision process.
\newblock {\em IEEE Transactions on Aerospace and Electronic Systems}, 49(4),
  2013.

\bibitem{Temizer}
Temizer S., Kochenderfer M., Kaelbling L., Lozano-Perez T., and Kuchar J.
\newblock Collision avoidance for unmanned aircraft using markov decision
  processes.
\newblock In {\em AIAA Guidance, Navigation, and Control Conference, Guidance,
  Navigation, and Control and Co-located Conferences}, 2010.

\bibitem{Kang}
K.~Thar S.~W.~Kang and C.~S. Hong.
\newblock Unmanned aerial vehicle allocation and deep learning based content
  caching in wireless network.
\newblock In {\em 2020 International Conference on Information Networking
  (ICOIN)}, pages 793--796, 2020.

\bibitem{saad}
Walid Saad, Mehdi Bennis, Mohammad Mozaffari, and Xingqin Lin.
\newblock {\em Wireless Communications and Networking for Unmanned Aerial
  Vehicles}.
\newblock Cambridge University Press, 2020.

\bibitem{Oubbati2}
O.~{Sami Oubbati}, M.~{Atiquzzaman}, T.~{Ahamed Ahanger}, and A.~{Ibrahim}.
\newblock Softwarization of uav networks: A survey of applications and future
  trends.
\newblock {\em IEEE Access}, 8:98073--98125, 2020.

\bibitem{Butenko}
Carlos A.~Oliveira Sergiy~Butenko, Xiuzhen~Cheng and P.~M. Pardalos.
\newblock Recent developments in cooperative control and optimization.
\newblock {\em A New Heuristic for the Minimum Connected Dominating Set Problem
  on Ad Hoc Wireless Networks}, pages 61--73, 2004.

\bibitem{Shakeri30}
R.~{Shakeri}, M.~A. {Al-Garadi}, A.~{Badawy}, A.~{Mohamed}, T.~{Khattab}, A.~K.
  {Al-Ali}, K.~A. {Harras}, and M.~{Guizani}.
\newblock Design challenges of multi-uav systems in cyber-physical
  applications: A comprehensive survey and future directions.
\newblock {\em IEEE Communications Surveys Tutorials}, 21(4):3340--3385, 2019.

\bibitem{Shamsoshoara}
A.~Shamsoshoara, M.~Khaledi, F.~Afghah, A.~Razi, and J.~Ashdown.
\newblock Distributed cooperative spectrum sharing in uav networks using
  multi-agent reinforcement learning.
\newblock In {\em 2019 16th IEEE Annual Consumer Communications and Networking
  Conference (CCNC 2019)}, 2019.

\bibitem{SharmaSurvey}
A.~Sharma, P.~Vanjani, N.~Paliwal, C.~Basnayaka, D.~Jayakody, H.~Wang, and
  P.~Muthuchidambaranathan.
\newblock Communication and networking technologies for uavs: A survey.
\newblock {\em Journal of Network and Computer Applications}, 168(8):149–169,
  2020.

\bibitem{Sharma2}
R.K. Sharma and D.~Ghose.
\newblock Collision avoidance between uav clusters using swarm intelligence
  techniques.
\newblock {\em International Journal of Systems Science}, pages 521--538, 2009.

\bibitem{ShenYuFPLinQ}
K.~{Shen} and W.~{Yu}.
\newblock Fplinq: A cooperative spectrum sharing strategy for device-to-device
  communications.
\newblock In {\em 2017 IEEE International Symposium on Information Theory
  (ISIT)}, pages 2323--2327, 2017.

\bibitem{ShiRaz}
Q.~{Shi}, M.~{Razaviyayn}, Z.~{Luo}, and C.~{He}.
\newblock An iteratively weighted mmse approach to distributed sum-utility
  maximization for a mimo interfering broadcast channel.
\newblock {\em IEEE Transactions on Signal Processing}, 59(9):4331--4340, 2011.

\bibitem{Shyalika}
T.~Shyalika, C. amd~Silva and A.~Karunananda.
\newblock Reinforcement learning in dynamic task scheduling: A review.
\newblock {\em SN COMPUT. SCI. 1, 306}, 2020.

\bibitem{Suganthi}
S.~Suganthi and S.P Rajagopalan.
\newblock Multi-swarm particle swarm optimization for energy-effective
  clustering in wireless sensor networks.
\newblock {\em Wireless Pers Commun}, 94:2487–2497, 2017.

\bibitem{SunLearning}
H.~{Sun}, X.~{Chen}, Q.~{Shi}, M.~{Hong}, X.~{Fu}, and N.~D. {Sidiropoulos}.
\newblock Learning to optimize: Training deep neural networks for interference
  management.
\newblock {\em IEEE Transactions on Signal Processing}, 66(20):5438--5453,
  2018.

\bibitem{tahir}
Anam Tahir, Jari Boeling, Mohammad-Hashem Haghbayan, Hannu T.Toivonen, and Juha
  Plosila.
\newblock Swarms of unmanned aerial vehicles — a survey.
\newblock {\em Journal of Industrial Information Integration}, 16, 2019.

\bibitem{AlladiBlockchain}
Nishad~Sahu Tejasvi~Alladi, Vinay~Chamola and Mohsen Guizani.
\newblock Applications of blockchain in unmanned aerial vehicles: A review.
\newblock {\em Vehicular Communications}, 23, 2020.

\bibitem{TolbaRL}
Amr Tolba and Abdulaziz Alarifi.
\newblock Optimizing the network energy of cloud assisted internet of things by
  using the adaptive neural learning approach in wireless sensor networks.
\newblock {\em Computers in Industry}, 106:133--141, 2019.

\bibitem{Tsai}
Yao-Hong Tsai.
\newblock Vision-based collision avoidance for unmanned aerial vehicles by
  recurrent neural networks.
\newblock {\em World Academy of Science, Engineering and Technology,
  International Journal of Computer and Information Engineering}, 13(4), 2019.

\bibitem{Ullaha}
Zaib Ullaha, Fadi~Al Turjmancd, Uzair Moatasimb, Leonardo Mostardaa, and
  Roberto Gagliardia.
\newblock Uavs joint optimization problems and machine learning to improve the
  5g and beyond communication.
\newblock {\em Computer Networks}, 182, 2020.

\bibitem{Venturini}
Federico Venturini, Federico Mason, Francesco Pase, Federico Chiariotti,
  Alberto Testolin, Andrea Zanella, and Michele Zorzi.
\newblock Distributed reinforcement learning for flexible uav swarm control
  with transfer learning capabilities.
\newblock In {\em Proceedings of the 6th ACM Workshop on Micro Aerial Vehicle
  Networks, Systems, and Applications}, 2020.

\bibitem{WangZhao}
H.~{Wang}, H.~{Zhao}, J.~{Zhang}, D.~{Ma}, J.~{Li}, and J.~{Wei}.
\newblock Survey on unmanned aerial vehicle networks: A cyber physical system
  perspective.
\newblock {\em IEEE Communications Surveys Tutorials}, 22(2):1027--1070, 2020.

\bibitem{WangGlowworm}
J.~Wang, B.~Li Y.-Q.~Cao, S.-Y. Lee, and J.-U. Kim.
\newblock A glowworm swarm optimization based clustering algorithm with mobile
  sink support for wireless sensor networks.
\newblock {\em Internet Technol. J.}, 16(5):825--832, 2015.

\bibitem{WANG2018339}
Zhu Wang, Li~Liu, Teng Long, and Yonglu Wen.
\newblock Multi-uav reconnaissance task allocation for heterogeneous targets
  using an opposition-based genetic algorithm with double-chromosome encoding.
\newblock {\em Chinese Journal of Aeronautics}, 31(2):339 -- 350, 2018.

\bibitem{Wu5G}
Qingqing Wu, Jie Xu, Yong Zeng, Derrick Wing~Kwan Ng, Naofal Al-Dhahir, Robert
  Schober, and A.~Lee Swindlehurst.
\newblock 5g-and-beyond networks with uavs: From communications to sensing and
  intelligence.
\newblock {\em arXiv:2010.09317}, 2020.

\bibitem{WuLi}
Z.~{Wu}, J.~{Li}, J.~{Zuo}, and S.~{Li}.
\newblock Path planning of uavs based on collision probability and kalman
  filter.
\newblock {\em IEEE Access}, 6:34237--34245, 2018.

\bibitem{LiuLiu}
Y.~Chen X.~Liu, Y.~Liu and L.~Hanzo.
\newblock Trajectory design and power control for multi-uav assisted wireless
  networks: A machine learning approach.
\newblock {\em IEEE Transactions on Vehicular Technology}, 68(8):7957--7969,
  2019.

\bibitem{XuDistributed}
H.~{Xu} and L.~R.~G. {Carrillo}.
\newblock Distributed near optimal flocking control for multiple unmanned
  aircraft systems.
\newblock In {\em 2015 International Conference on Unmanned Aircraft Systems
  (ICUAS)}, pages 879--885, 2015.

\bibitem{Cao}
L.~Zhang Y.~Cao and Y.~Liang.
\newblock Deep reinforcement learning for channel and power allocation in
  uav-enabled iot systems.
\newblock In {\em 2019 IEEE Global Communications Conference (GLOBECOM)}, pages
  1--6, 2019.

\bibitem{Yu}
K~Fang Y~Yu, L~Ru.
\newblock Bio-inspired mobility prediction clustering algorithm for ad hoc uav
  networks.
\newblock {\em Engineering Letters}, 2016.

\bibitem{YANG2019140}
Jun Yang, Xinghui You, Gaoxiang Wu, Mohammad~Mehedi Hassan, Ahmad Almogren, and
  Joze Guna.
\newblock Application of reinforcement learning in uav cluster task scheduling.
\newblock {\em Future Generation Computer Systems}, 95:140 -- 148, 2019.

\bibitem{YangPan}
Z.~{Yang}, C.~{Pan}, K.~{Wang}, and M.~{Shikh-Bahaei}.
\newblock Energy efficient resource allocation in uav-enabled mobile edge
  computing networks.
\newblock {\em IEEE Transactions on Wireless Communications}, 18(9):4576--4589,
  2019.

\bibitem{Ye20}
Fang Ye, Jie Chen, Yuan Tian, and Tao Jiang.
\newblock Cooperative multiple task assignment of heterogeneous uavs using a
  modified genetic algorithm with multi-type-gene chromosome encoding strategy.
\newblock {\em J Intell Robot Syst}, page 615–627, 2020.

\bibitem{Yuan17}
X.~Elhoseny Yuan, M.~El-Minir, and H.K. et~al.
\newblock A genetic algorithm-based, dynamic clustering method towards improved
  wsn longevity.
\newblock {\em J Netw Syst Manage}, 25:21–46, 2017.

\bibitem{Zeng}
Tengchan Zeng, Omid Semiari, Mohammad Mozaffari, Mingzhe Chen, Walid Saad, and
  Mehdi Bennis.
\newblock Federated learning in the sky: Joint power allocation and scheduling
  with uav swarms.
\newblock {\em arXiv:2002.08196}, 2020.

\bibitem{ZengNavigation}
Yong Zeng, Xiaoli Xu, Shi Jin, and Rui Zhang.
\newblock Simultaneous navigation and radio mapping for cellular-connected uav
  with deep reinforcement learning.
\newblock {\em arXiv:2003.07574}, 2020.

\bibitem{Zhang}
Jing Zhang, Minhao Lou, Lin Xiang, and Long Hu.
\newblock Power cognition: Enabling intelligent energy harvesting and resource
  allocation for solar-powered uavs.
\newblock {\em Future Generation Computer Systems}, 2019.

\bibitem{ZhangDeployment}
Qianqian Zhang, Mohammad Mozaffari, Walid Saad, mérouane Debbah, Mehdi Bennis,
  and Mérouane Debbah.
\newblock Machine learning for predictive on-demand deployment of uavs for
  wireless communications.
\newblock In {\em 2018 IEEE Global Communications Conference}, 2018.

\bibitem{ZhangLi}
W.~{Zhang}, L.~{Li}, N.~{Zhang}, T.~{Han}, and S.~{Wang}.
\newblock Air-ground integrated mobile edge networks: A survey.
\newblock {\em IEEE Access}, 8:125998--126018, 2020.

\bibitem{ZhangLi2}
Z.~{Zhang}, L.~{Li}, X.~{Liu}, W.~{Liang}, and Z.~{Han}.
\newblock Matching-based resource allocation and distributed power control
  using mean field game in the noma-based uav networks.
\newblock In {\em 2018 Asia-Pacific Signal and Information Processing
  Association Annual Summit and Conference (APSIPA ASC)}, pages 420--426, 2018.

\bibitem{Zhao2019588}
Xinyi Zhao, Qun Zong, Bailing Tian, Boyuan Zhang, and Ming You.
\newblock Fast task allocation for heterogeneous unmanned aerial vehicles
  through reinforcement learning.
\newblock {\em Aerospace Science and Technology}, 92:588--594, 2019.

\end{thebibliography}
\end{document}